\documentclass[5p,twocolumn]{elsarticle} 
\usepackage{amsmath,amssymb,amsfonts} 
\usepackage{mathtools, nccmath}
\usepackage{algorithmic}
\usepackage[ruled,vlined]{algorithm2e}
\usepackage{graphicx}
\usepackage{subcaption}
\usepackage{framed}
\usepackage[table,xcdraw]{xcolor}
\usepackage{textcomp}
\usepackage{hyperref}
\usepackage{natbib}
\usepackage{enumitem}
\usepackage{multirow}
\usepackage{xcolor}
\usepackage{amsmath}
\usepackage{tabularx}
\DeclareMathOperator*{\payofff}{\mathit{payoff}}
\DeclareMathOperator*{\KF}{\mathit{kf}}
\makeatletter

\newcommand{\Rmnum}[1]{\expandafter\@slowromancap\romannumeral #1@}
\makeatother
\usepackage{makecell}

\newcommand{\commentout}[1]{}

\newcommand{\real}{\mathbb{R}}

\newcommand{\network}{{N}}

\newcommand{\videoframe}{\alpha}

\newcommand{\inputdomain}{{\tt X}}
\newcommand{\outputdomain}{{\tt Y}}

\newcommand{\dist}{{\cal D}}

\newcommand{\POsquare}{\{PO\}\textsuperscript{2}}

\newcommand{\matr}[1]{\mathbf{#1}}

\newcommand{\payoff}{\pi}
\newcommand{\partialorder}{\beta}

\begin{document}

\begin{frontmatter}
\title{\LARGE \bf Formal Verification of Robustness and Resilience of \\Learning-Enabled State Estimation Systems\tnoteref{label1}}

\author[1,2]{Wei Huang\corref{corrr1}}
\ead{huangwei@pmlabs.com.cn}

\author[2]{Yifan Zhou}
\ead{Yifan.Zhou@liverpool.ac.uk}

\author[2]{Gaojie Jin}
\ead{g.jin3@liverpool.ac.uk}

\author[3]{Youcheng Sun}
\ead{youcheng.sun@manchester.ac.uk}

\author[5]{Jie Meng}
\ead{j.meng@lboro.ac.uk}

\author[6]{Fan Zhang}
\ead{17034203@qq.com}

\author[2]{Xiaowei Huang}
\ead{xiaowei.huang@liverpool.ac.uk}

\cortext[corrr1]{Corresponding author}

\affiliation[1]{organization={Purple Mountain Laboratories},
country={China}}

\affiliation[2]{organization={University of Liverpool},
country={UK}}

\affiliation[3]{organization={University of Manchester},
country={UK}}

\affiliation[5]{organization={Loughborough University London},
country={UK}}

\affiliation[6]{organization={National Digital Switching System Engineering and Technological Research Center},
country={China}}

\tnotetext[label1]{A preprint has been previously published in \cite{huang2020formal}. Part of this paper appeared in \cite{SZMSH2020,9340720}.}

\begin{abstract}
This paper presents a formal verification guided approach for a principled design and implementation of robust and resilient learning-enabled systems. We focus on 
learning-enabled state estimation systems (LE-SESs), which have been widely used in robotics applications to determine the current state (e.g., location, speed, direction, etc.) of a complex system. The LE-SESs are networked systems, composed of a set of connected components including: Bayes filters for state estimation, and neural networks for processing sensory input. We study LE-SESs from the perspective of formal verification, which determines the satisfiabilty of a system model against the specified properties. Over LE-SESs, we investigate two key properties --  robustness and resilience -- and provide their formal definitions. To enable formal verification, we reduce the LE-SESs to a novel class of labelled transition systems, named \POsquare-LTS in the paper, and formally express the properties as constrained optimisation objectives. We prove that the verification problems are NP-complete. Based on \POsquare-LTS and the optimisation objectives, practical verification algorithms are developed to check the satisfiability of the properties on the LE-SESs. As a major case study, we interrogate a real-world dynamic tracking system which uses a single Kalman Filter (KF) -- a special case of Bayes filter -- to localise and track a ground vehicle. Its perception system, based on convolutional neural networks, processes a high-resolution Wide Area Motion Imagery (WAMI) data stream. Experimental results show that our algorithms can not only verify the properties of the WAMI tracking system but also provide representative examples, the latter of which inspired us to take an enhanced LE-SESs design where runtime monitors or joint-KFs are required. Experimental results confirm the improvement in the robustness of the enhanced design.

\end{abstract}
\end{frontmatter}

\section{Introduction}
Autonomous systems, marked by their advanced intelligence, can autonomously make decisions based on both their internal state and external environment perception. These systems are often constructed by integrating diverse components, forming a networked system \cite{Sifakis2018}. A prominent category of such autonomous systems, widely employed in the field of robotics, is referred to as state estimation systems (SESs). SESs play a vital role in determining the real-time state of dynamic systems, such as spacecraft and ground vehicles, including aspects like their location, speed, and direction. In the realm of robotics, SESs find applications in tasks such as localization \cite{papadopoulos2010cooperative}, tracking \cite{gordon1995bayesian}, and control \cite{ZM2019}.

Failures within SESs can result from various factors, including sensor-actuator faults, Denial-of-Service (DoS) attacks, equipment aging-related uncertainties, measurement inaccuracies, and more. While significant research has been devoted to developing novel control algorithms to address these issues, such as adaptive neural finite-time resilient dynamic surface control \cite{song2023finite}, point-to-point iterative learning control for constrained systems \cite{zhou2022robust}, and switching-like event-triggered approaches for reaction-diffusion neural network systems \cite{song2023switching}, there has been limited investigation into failures associated with Deep Neural Network (DNN) components in SESs. These integrated systems are often termed learning-enabled SESs (LE-SESs), and their significance has grown with the increasing adoption of DNN components in robotics applications due to their superior predictive accuracy \cite{he2017adaptive}.

In LE-SESs, neural networks are commonly used to process sensory inputs from various sensors. For instance, Convolutional Neural Networks (CNNs) are frequently employed to handle image inputs, providing essential data for state estimation. However, neural networks have exhibited robustness vulnerability, as they can be manipulated by subtle yet valid alterations to input, resulting in incorrect classification outputs, a phenomenon known as adversarial attacks \cite{szegedy2014intriguing}. Various strategies have been developed to enhance the robustness of neural networks, including techniques for mitigating adversarial attacks \cite{PMWJS2016,CW2016,jin2022enhancing}, formal verification methods  \cite{HKWW2017,wu2018game,ruan2018global,LLYCH2018}, and coverage-guided testing \cite{sun2018concolic,huang2021coverage,xie2022npc}. These efforts collectively contribute to assessing the trustworthiness of systems incorporating neural networks, reflecting the confidence that such systems will produce accurate outputs for a given input. For an extensive overview of this topic, a recent survey is available in \cite{huang2020survey}.

While in the context of neural networks, the concepts of robustness and resilience are closely related. It is uncertain whether this relationship holds for LE-SESs or, more broadly, networked systems with learning components. Prior research has used both terms, "robustness" and "resilience" to describe the ability of SESs to withstand component failures or input disturbances \cite{zhou2022robust,song2023finite,song2023switching}. This paper will demonstrate that, \textbf{for LE-SESs, subtle yet significant distinctions exist between robustness and resilience}. In general terms, robustness signifies a system's ability to consistently deliver its 'expected' functionality, even in the presence of disturbances to the input. Contrastively, resilience refers to the system's capacity to withstand and recover from challenging conditions, which may involve internal failures and external shocks, all while preserving or resuming a portion, if not all, of its designated functionality. Based on this perspective, the paper will propose formal definitions of robustness and resilience within the context of LE-SESs.

In the previous study \cite{SZMSH2020}, it was observed that the LE-SESs exhibit signs of robustness by compensating to some extent against adversarial attacks on its neural network component and signs of resilience by recovering from deviations. Additionally, a formal verification methods for assessing the robustness of LE-SESs was proposed in \cite{9340720}. In this paper, we take a step further by systematically studying the robustness and resilience of real-world LE-SESs. We apply formal verification techniques to demonstrate that a system is correct against all possible risks over a given specification, along with the formal model of the system, and it provides counter-examples when it cannot meet these criteria. This approach is crucial for identifying risks before deploying safety-critical applications.

Technically, we first formalise an LE-SES as a novel labelled transition system which has components for payoffs and partial order relations (i.e. relations that are reflexive, asymmetric and transitive). The labelled transition system is named \POsquare-LTS in the paper. Specifically, every transition is attached with a payoff, and for every state there is a partial order relation between its out-going transitions from the same state. Second, we show that the verification of the  properties -- both robustness and resilience -- on such a system can be reduced into a constrained optimisation problem. Third, we prove that the verification problem is NP-complete on \POsquare-LTS. Fourth, to enable practical verification, we develop an automated verification algorithm.

As a major case study, we work with a real-world dynamic tracking system \cite{ZM2019}, which detects and tracks ground vehicles over the high-resolution Wide Area Motion Imagery (WAMI) data stream, named \emph{WAMI tracking system} in this paper. The system is composed of two major components: a state estimation unit and a perceptional unit. The perceptional unit includes multiple, networked CNNs, and the state estimation unit includes one or multiple Kalman filters, which are a special case of Bayes filter. We apply the developed algorithm to the WAMI tracking system to analyse both robustness and resilience, in order to understand whether the system can function well when subject to adversarial attacks on the perceptional neural network components. 

The formal verification approach leads to a guided design of the LE-SESs. 
As the first design, we use a single Kalman filter to interact with the perceptional unit, and our experimental results show that the LE-SES performs very well in a tracking task, when there is no attack on the perceptional unit. However, it may perform less well in some cases when the perceptional unit is under adversarial attack. The returned counterexamples from our verification algorithms indicate that we may improve the safety performance of the system by adopting a better design. Therefore, a second, improved design -- with joint-KFs to associate observations and/or a runtime monitor -- is taken. Joint-KFs increase the capability of the system in dealing with internal and external uncertainties, and a runtime monitor can reduce some potential risks. We show that in the resulting LE-SES, the robustness is improved, without compromising the precision of the tracking.

\begin{framed}

The main contributions of this paper are as follows.
\begin{enumerate}
    \item \emph{Robustness vs resilience:} This paper pioneers in aligning the definitions of robustness and resilience in LE-SESs with those applied in traditional high-integrity computing. Their similarity and difference are examined both in theory (Section~\ref{sec:properties}) and in experimental evaluation (Section~\ref{sec:experiments}).
    \item \emph{Formal guarantee:} The robustness and resilience of the LE-SES is guaranteed by a novel formal verification technique (Sections~\ref{opti_prob},\ref{sec:properties},\ref{sec:algorithms}).
    \item \emph{Robust and resilient LE-SES design:} This paper proposes a principled and detailed design of robust and resilient learning-enabled state estimation systems (Section~\ref{sec:design2}).
\end{enumerate}
\end{framed}

In summary, the paper's organization is structured as follows: We begin by presenting the foundational concepts of neural networks and LE-SESs with the use of Bayes (or Kalman) filters in the next section. The paper then delves into the reduction of the LE-SES system to the \POsquare-LTS in Section~\ref{opti_prob}. Section~\ref{sec:properties} offers a methodological discussion on distinguishing between robustness and resilience, along with their formalization as optimization objectives. The automated verification algorithm is introduced in Section~\ref{sec:algorithms}. The case study of LESs, focusing on the WAMI-tracking system, is presented in Section~\ref{sec:WAMI_intro}. Section~\ref{sec:design2} outlines our improved design for the WAMI-tracking system, incorporating a runtime monitor and/or joint-KFs. Experimental results are shared in Section~\ref{sec:experiments}. Section~\ref{sec:discussion} provides insights into potential future directions. Lastly, we delve into related work in Section~\ref{sec:related} before concluding the paper in Section~\ref{sec:concl}.

\section{Preliminaries}\label{sec:preliimnaries}

\subsection{Convolutional Neural Networks}

Let $\inputdomain$ be the input domain and $\outputdomain$ be the set of labels. A neural network $\network:\inputdomain\rightarrow \dist(\outputdomain)$ can be seen as a function mapping from $\inputdomain$ to probabilistic distributions over $\outputdomain$. That is, $\network(x)$ is a probabilistic distribution, which assigns for each label $y\in \outputdomain$ a probability value $(\network(x))_y$. We let $f_\network:\inputdomain\rightarrow \outputdomain$ be a function such that for any $x\in \inputdomain$, 
$f_\network(x) = \arg\max_{y\in \outputdomain}\{(\network(x))_y\}$, i.e., $f_\network(x)$ returns the classification.

\subsection{Learning Enabled State Estimation}\label{sec:state-estimation}

We consider a time-series linear state estimation problem that is widely assumed in the context of object tracking. The process model is defined as follows.
\begin{equation} \label{equ:process_model1}
    \matr{s}_{k} =\matr{F} \cdot \matr{s}_{k-1} + \matr{\omega}_{k}
\end{equation}
where $\matr{s}_{k}$ is the state at time $k$, $\matr{F}$ is the transition matrix, $\matr{\omega}_{k}$ is a zero-mean Gaussian noise such that $\matr{\omega}_{k} \sim \mathcal{N}(0, \matr{Q})$, with $\matr{Q}$ being the covariance of the process noise. Usually, the states are not  observable and need to be determined indirectly by  measurement and reasoning. The measurement model is defined as:
\begin{equation} \label{equ: measuremnet_model}
    \matr{z}_{k} = \matr{H} \cdot \matr{s}_{k} + \matr{v}_{k}
\end{equation}
where $\matr{z}_{k}$ is the observation, $\matr{H}$ is the measurement matrix, $\matr{v}_{k}$ is a zero-mean Gaussian noise such that $\matr{v}_{k}\sim\mathcal{N}(0, \matr{R})$, and $\matr{R}$ is the covariance of the measurement noise.

Bayes filters have been used for reasoning about the observations, $\{\matr{z}_k\}$, with the goal of learning the underlying states $\{\matr{s}_k\}$. A Bayes filter maintains a pair of variables, $(\matr{s}_{k},\matr{P}_{k})$, over the time, denoting Gaussian estimate and Bayesian uncertainty, respectively. The basic procedure of a Bayes filter is to use a transition matrix, $\matr{F}_k$, to predict the current state, $(\hat{\matr{s}}_{k},\hat{\matr{P}}_{k})$, given the previous state, $(\matr{s}_{k-1},\matr{P}_{k-1})$. The prediction state can be updated into $(\matr{s}_{k},\matr{P}_{k})$ if a new observation, $\matr{z}_{k}$, is obtained. In the context of the aforementioned problem, this procedure is iterated for a number of time steps, and is always discrete-time, linear, but subject to noises.

We take the Kalman Filter (KF), one of the most widely used variants of Bayes filter, as an example to demonstrate the above procedure. Let $\matr{s}_0\in \real^n\sim {\cal N}(\hat{\matr{s}}_0,\hat{\matr{P}}_0)$ be the initial state, such that $\hat{\matr{s}}_0\in \real^n$ and $\hat{\matr{P}}_0\in \real^{n\times n}$ represent our knowledge about the initial estimate and the corresponding covariance matrix, respectively.

First, we perform the \textbf{state prediction} for $k\geq 1$:
\begin{equation}
\label{eq:kf_predict}
\begin{array}{lcl}
\hat{\matr{s}}_k & = & \matr{F}_k \matr{s}_{k-1} \\
\hat{\matr{P}}_{k} & = & \matr{F}_k \matr{P}_{k-1} \matr{F}_k^T + \matr{Q}_k
\end{array}
\end{equation}

Then, we can \textbf{update the filter}:
\begin{equation}
\label{eq:kf_update}
\begin{array}{lcl}
\matr{s}_{k} & = & \hat{\matr{s}}_{k} + \matr{K}_k \matr{y}_k \\ 
\matr{P}_{k} & = & (\matr{I} - \matr{K}_k \matr{H}_k)\hat{\matr{P}}_{k} 
\end{array}
\end{equation}
where
\begin{equation}\label{eq:kf_update2}
\begin{array}{lcl}
\matr{y}_k &  = & \matr{z}_k - \matr{H}_k \hat{\matr{s}}_{k} \\ 
\matr{S}_k & = & \matr{H}_k \hat{\matr{P}}_{k} \matr{H}_k^T + \matr{R}_k \\
\matr{K}_k & = & \hat{\matr{P}}_{k} \matr{H}_k^T \matr{S}_k^{-1}
\end{array}
\end{equation}

Intuitively, $\matr{y}_k$ is usually called ``innovation'' in signal processing and represents the difference between the real observation and the predicted observation, $\matr{S}_k$ is the covariance matrix of this innovation, and $\matr{K}_k$ is the Kalman gain, representing the relative importance of innovation $\matr{y}_k$ with respect to the predicted estimate $\hat{\matr{s}}_k$.

In a neural network enabled state estimation, a perception system  -- which may include multiple CNNs -- will provide a set of candidate observations $Z_k$, any of which can be chosen as the new observation $\matr{z}_k$. From the perspective of robotics, $Z_k$ includes a set of possible states of the robot, measured by (possibly several different) sensors at time $k$. These measurements are imprecise, and are subject to noise from both the environment (epistemic uncertainty) and the imprecision of sensors (aleatory uncertainty).

\section{Reduction of LE-SESs to Labelled Transition Systems} \label{opti_prob}

Formal verification requires a formal model of the system so that all possible behaviors of the system model can be explored to understand the existence of incorrect behavior. In this section, we will reduce the LE-SESs to a novel class of labeled transition systems, which serve as a formal model, ensuring the preservation of all safety-related behaviors. In the next few sections, we will discuss the formalization of the properties and the automated verification algorithms, respectively.

\subsection{Threat Model of Adversarial Attack on Perception System} \label{percep}
A neural network based perception system provides the observation $\matr{z}_k$. Let $x(\matr{z}_k)\in\real^{d_1\times d_2}$ be an image covering the observations $\matr{z}_k$, a neural network function $f_\network:\real^{d_1\times d_2}\rightarrow \{0,1\}$ maps $x(\matr{z}_k)$ into a Boolean value, $f_\network(x(\matr{z}_k))$, representing whether or not a observation is present. There are two types of erroneous detection: (1) a wrong classification prediction of the image $x(\matr{z}_k)$, and (2) a bias measurement of observations within $x(\matr{z}_k)$. We focus on the former since the LESs has a comprehensive mechanism to prevent the occurrence of the latter through novel control algorithm\cite{song2023finite,song2023switching}. 

\begin{figure}
    \centering
    \includegraphics[width=1\linewidth]{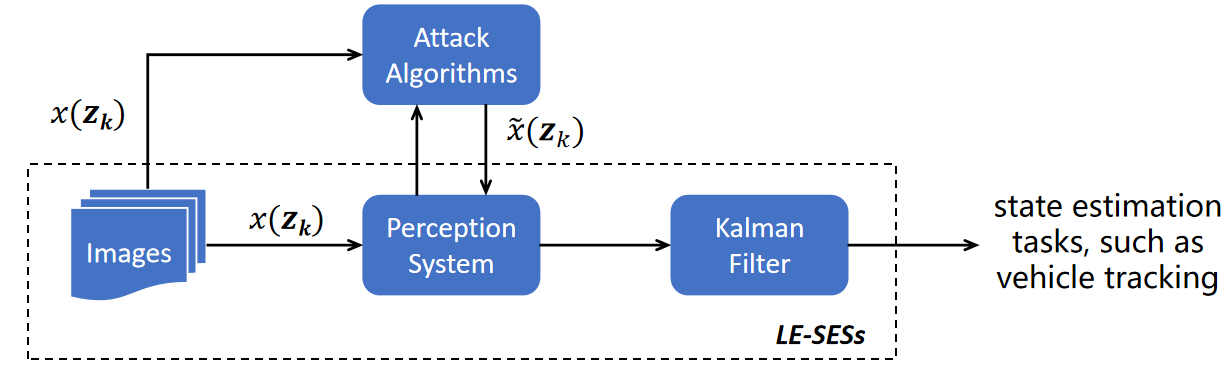}
    \caption{The Workflow of Attack on LE-SESs.}
    \label{fig:verif_framework}
\end{figure}

The threat model of an adversary is summarised as in Figure~\ref{fig:verif_framework}. Assuming that $f_\network(x(\matr{z}_k))=1$. An adversary must compute  
another input $\widetilde{x}(\matr{z}_k)$ which requires a  payoff and has a different classification, i.e., $f_\network(\widetilde{x}(\matr{z}_k))=0$. Without loss of generality, the $\mathit{payoff}$ is measured with the norm-distance from $\widetilde{x}(\matr{z}_k)$ to its original image $x(\matr{z}_k)$, or formally
\begin{equation}
    \payoff = ||\widetilde{x}(\matr{z}_k)-x(\matr{z}_k)||_p
\end{equation}

To deviate from an input image $x(\matr{z}_k)$ to its adversarial input $\widetilde{x}(\matr{z}_k)$, a large body of adversarial example generation algorithms and adversarial test case generation algorithms are available. Given a neural network $\network$ and an input $x$, an adversarial algorithm $A$ produces an adversarial example $A(\network,x)$ such that $f_\network(A(\network,x))\neq f_\network(x)$. On the other hand, for test case generation, an algorithm $A$ produces a set of test cases $A(\network,x)$, among which the optimal adversarial test case is such that $\arg\min_{\widetilde{x}\in A(\network,x),f_\network(\widetilde{x})\neq f_\network(x)}||\widetilde{x} - x||_p$. We note that a more advanced adversarial algorithm can produce a more accurate measure of attack payoff, which in turn can influence the verification results. In our experiments, we employ the two most popular and powerful algorithms:

\begin{itemize}
\item DeepFool \cite{moosavi2016deepfool}, which finds an adversarial example $\widetilde{x}$ by projecting $x$ onto the nearest decision boundary. The projection process is iterative  because the decision boundary is non-linear.
\item DeepConcolic \cite{sun2018concolic}, which generates a test suite by applying combined symbolic execution and concrete execution, guided by adapted MC/DC metrics for neural networks \cite{sun2018testing}. 
\end{itemize}

We denote by $\payoff = \payofff(A,\network,x)$, the payoff that an adversarial algorithm $A$ needs to compute for an adversarial example from $x$ and $\network$. Furthermore, we assume that the adversary can observe the parameters of the Bayes filter, for example, $\matr{H}_k, \matr{F}_k$, $\matr{Q}_k, \matr{R}_k$ of the Kalman filter.

\subsection{\texorpdfstring{\POsquare}-Labelled Transition Systems}

Let $Prop$ be a set of atomic propositions. A payoff and partially-ordered label transition system, or \POsquare-LTS, is a tuple $M=(Q,\rho_0,\KF,L,\payoff,\partialorder)$, where $Q$ is a set of states, $\rho_0\in Q$ is an initial state, $\KF\subseteq Q\times Q$ is a transition relation, $L:Q\rightarrow 2^{Prop}$ is a labelling function, $\payoff:Q\times Q\rightarrow \real^+$ is a payoff function assigning every transition a non-negative real number, and $\partialorder:\KF\rightarrow \KF$ is a partial order relation between out-going transitions from the same state.

\subsection{Reduction of LE-SESs to \texorpdfstring{\POsquare}-LTS}

We model a neural network enabled state estimation system as a \POsquare-LTS. A brief summary of some key notations in this paper is provided in Table~\ref{tab:summary}. We let each pair $(\matr{s}_k,\matr{P}_k)$ be a state, and use the transition relation $\KF$ to model the transformation from a pair to another pair in a Bayes filter. Assume we have the initial state $\rho_0 =(\matr{s}_0,\matr{P}_0)$. 
From a state $\rho_{k-1}=(\matr{s}_{k-1},\matr{P}_{k-1})$ and a set $Z_k$ of candidate observations, we have one transition $(\rho_{k-1},\rho_{k})$ for each $\matr{z}\in Z_k$, where $\rho_{k}=(\matr{s}_{k},\matr{P}_{k})$ can be computed with Equations (\ref{eq:kf_predict})-(\ref{eq:kf_update2}) by having $\matr{z}_k$ as the new observation. Subsequently, for each transition $(\rho_{k-1},\rho_{k})$, its associated payoff $\payoff(\rho_{k-1},\rho_{k})$ is denoted by $\payofff(A,\network,x(\matr{z}_k))$, i.e., the payoff that the adversary uses the algorithm $A$ to manipulate the image covering the observation $x(\matr{z}_k)$ into another image on which the neural network $\network$ believes there exists no observation. 

\begin{table}[ht]
\centering
\resizebox{\linewidth}{!}{
\begin{tabular}{c|c}
 \hline
 \textbf{Notations}                 &       \textbf{Description} \\\hline
 $Z_k$                &      a set of candidate observations\\ \hline
 $x(\matr{z}_k)$                      &       an $d_1$$\times$$d_2$ image covering observation $\matr{z}_k$\\ \hline
 $f_N$                              &       neural network function\\ \hline
 $\payoff = \payofff(A,\network,x)$           &       \makecell{payoff for algorithm $A$ computing\\ an adversarial example from $x$ and $\network$}\\ \hline
 $\rho_k=(\matr{s}_k,\matr{P}_k)$      &       \makecell{a state at step $k$, consisting of\\ estimate and covariance matrix}\\ \hline
 $\matr{s}_k$, $\matr{P}_k$ and $\matr{z}_k$   &   \makecell{estimate, covariance matrix \\ and observation for transition $(\rho_{k-1},\rho_{k})$}\\ \hline
 $\rho$                             &       a path of consecutive states $\rho_l...\rho_u$ \\
 \hline
\end{tabular}}
\caption{A Summary of Notations Used}
\label{tab:summary}
\end{table}

For two transitions $(\rho_{k-1},\rho_k^{1})$  and  $(\rho_{k-1},\rho_k^{2})$ from the same state $\rho_{k-1}$, we say that they have a partial order relation, written as  $(\rho_{k-1},\rho_k^{1}) \prec (\rho_{k-1},\rho_k^{2})$, if making $\matr{z}_k^{2}$ the new observation requires the adversary to deceive the network $\network$ into misclassifying $x(\matr{z}_k^1)$. For example, in the case study of WAMI tracking (refer to Section~\ref{sec:WAMI_intro}), according to Equation (\ref{equ:WAMIobservation}), the condition means that $||\matr{z}_k^2-\matr{H}_k\hat{\matr{s}}_k||_p > ||\matr{z}_k^1-\matr{H}_k\hat{\matr{s}}_k||_p$, where $\matr{H}_k\hat{\matr{s}}_k$ is the predicted location.

\begin{figure}
    \centering
    \includegraphics[width=1\linewidth]{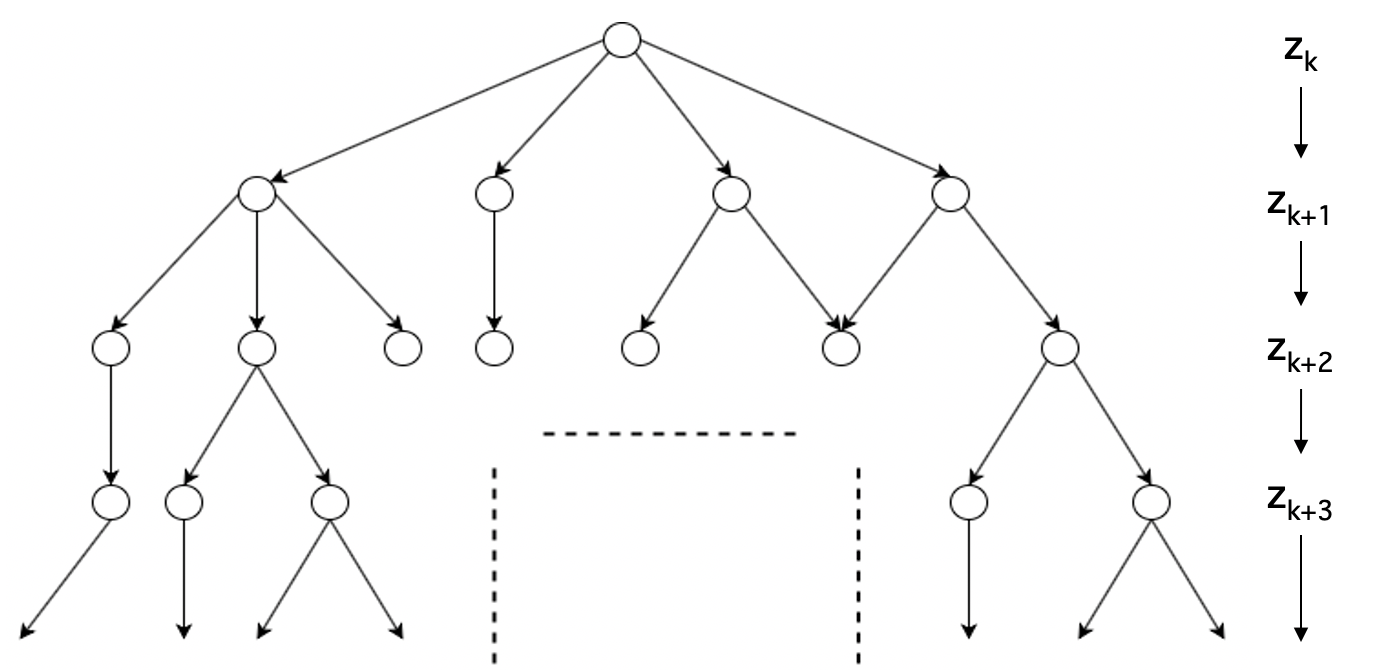}
    \caption{Tree diagram of an unfolding \POsquare-LTS}
    \label{fig:tree_diag}
\end{figure}

Figure~\ref{fig:tree_diag} depicts a tree diagram for the unfolding of a labelled transition system. The root node on top represents the initial state $\rho_0$. Each layer comprises all possible states of $\rho_k=(\matr{s}_{k},\matr{P}_{k})$ at step $k$, with $\matr{s}_{k}$ being one possible estimate, and $\matr{P}_{k}$ the covariance matrix. Each transition connects a state $\rho_{k-1}$ at step $k$$-$$1$ to $\rho_{k}$ at step $k$. $\dots,\matr{z}_k,\matr{z}_{k+1},\matr{z}_{k+2},\dots$ are observations at each step.

Given a \POsquare-LTS $M$, we define a path $\rho$ as a sequence of consecutive states $\rho_l...\rho_u$, and $\matr{z}_l...\matr{z}_u$ as a sequence of corresponding observed location for $0\le l<u$, where $l$ and $u$ are the starting and ending time under attack consideration, respectively.

\section{Property Specification: Robustness and Resilience}\label{sec:properties}

Formal verification determines whether a specification $\phi$ holds on a given LTS $M$ \cite{clarkebook}. Usually, a logic language, such as CTL, LTL, or PCTL, is used to formally express the specification  $\phi$, and for learning-enabled systems, there are recent work adapting these logic languages, such as \cite{10.1007/978-3-031-21222-2_8,BENSALEM2024100941}. In this paper, to suit our needs, we let the specification $\phi$ be a constrained optimisation objective; and so verification is undertaken in two steps: 

\begin{enumerate}
    \item determine whether, given $M$ and $\phi$, there is a solution to the constrained optimisation problem. If the answer is affirmative, an optimal solution $sol_{opt}(M,\phi)$ is returned. 
    \item compare $sol_{opt}(M,\phi)$ with a pre-specified threshold $\theta$. If $sol_{opt}(M,\phi) >  \theta$ then we say that the property $\phi$ holds on the model $M$ with respect to the threshold $\theta$. Otherwise, it fails. 
\end{enumerate} 
Note that, 
we always take a minimisation objective in the first step. Since the optimisation is to find the minimal answer, in the second step, if $sol_{opt}(M,\phi) > \theta$, we cannot have a better -- in terms of a smaller value -- solution for the threshold $\theta$. \textbf{\emph{Intuitively, it is a guarantee that no attacker can succeed with less cost than $\theta$, and the system is hence safe against the property.}} The above procedure can be easily adapted if we work with maximisation objectives. 

Before proceeding to the formal definition of the robustness and resilience properties, we need several notations. First, we consider the measure for the loss of precision. Let $\rho$ be an original path that has not suffered from an attack. The other path $\widetilde{\rho}$ is obtained after an attack on $\rho$. For the LE-SESs, we define their distance at time $k$ as 
\begin{equation}\label{eq:monitor-uncertainty2}
dist(\rho_k,\widetilde{\rho}_k)= || \matr{s}_k-\widetilde{\matr{s}}_k||_p
\end{equation} 
which is the $L_p$-norm difference between two states $\matr{s}_k$ and $\widetilde{\matr{s}}_k$.

Moreover, let $(\widetilde{\rho}_{k-1},\widetilde{\rho}_{k})$ be a transition on an attacked path $\widetilde{\rho}$, and so we have 
\begin{equation}
    \label{eq:combined-payoff}
    \varphi(\widetilde{\rho}_{k-1},\widetilde{\rho}_{k}) = \sum_{(\widetilde{\rho}_{k-1},\widetilde{\rho}_{k}^{\diamond})\prec (\widetilde{\rho}_{k-1},\widetilde{\rho}_{k})} \payoff(\widetilde{\rho}_{k-1},\widetilde{\rho}_{k}^{\diamond})
\end{equation}
as the \emph{combined payoffs} that are required to implement the transition $(\widetilde{\rho}_{k-1},\widetilde{\rho}_{k})$. Intuitively, it requires that all the payoffs of the transitions $(\widetilde{\rho}_{k-1},\widetilde{\rho}_{k}^{\diamond})$, which are partially ordered by the envisaged transition $(\widetilde{\rho}_{k-1},\widetilde{\rho}_{k})$, are counted. In the LE-SESs, this means that the attack results in misclassification of all the images $x(\widetilde{\matr{z}}_{k}^{\diamond})$ such that no observation  $\widetilde{\matr{z}}_{k}^{\diamond}$ is closer to the prediction $\matr{F}_k\widetilde{\matr{s}}_{k-1}$ than $\widetilde{\matr{z}}_{k}$.

\subsection{Definition of Robustness}

Robustness is a concept that has been studied in many fields such as psychology \cite{ruscio2008probability}, biomedical analysis \cite{vander2001guidance}, and chemical analysis \cite{gonze2002biochemical}. Here, we adopt the general definition of robustness as used in the field of artificial intelligence (we later discuss the difference between this and the definition applied in software engineering):

\begin{framed}
Robustness is an enforced measure to represent a system's ability to \emph{consistently} deliver its \emph{expected} functionality by accommodating disturbances to the input. 
\end{framed}

In LE-SESs, we measure the quality of the system maintaining its expected functionality under attack on a given scenario with the distance between two paths: its original path and the attacked path. 
Formally, given a path $\rho$ and an attacker, it is to consider the minimal perturbation to the input that can lead to malfunction. Intuitively, the larger the amount of perturbations a system can tolerate, the more robust it is. Let
\begin{equation}\label{eq:monitor-uncertainty}
dist^{0,e}(\rho,\widetilde{\rho})=\sum_{k=0}^e dist(\rho_k,\widetilde{\rho}_k)
\end{equation} 
be the accumulated distance, between the original path $\rho$ and the attacked path $\widetilde{\rho}$, from the start $k = 0$ to the end $k = e$.

Moreover, we measure the disturbances to an LE-SES as the perturbation to its imagery input. Formally, we let 
\begin{equation}\label{equ:varphi}
    \varphi^{l,u}(\widetilde{\rho}) = \sum_{k=l+1}^u\varphi(\widetilde{\rho}_{k-1},\widetilde{\rho}_{k})
\end{equation}
be the accumulated combined payoff for the attacked path $\widetilde{\rho}$ between time steps $l$ and $u$, such that $l \geq 0$ and $u \leq e$.
When $\varphi^{0,e}(\widetilde{\rho})=0$, there is no perturbation and $\widetilde{\rho}$ is the original track $\rho$.

Finally, we have the following optimisation objective for robustness:  
\begin{equation}
    \begin{array}{rl}
        \underset{\widetilde{\rho}}{\textbf{minimize}} & \varphi^{l,u}(\widetilde{\rho}) \\
        \textbf{subject to} & dist^{0,e}(\rho,\widetilde{\rho}) > \epsilon_{robustness} \\
    \end{array}
    \label{equ:robustnessgeneral}
\end{equation}
Basically, $\varphi^{l,u}(\widetilde{\rho})$ represents the amount of perturbation to the input, while the malfunctioning of system is formulated as $dist^{0,e}(\rho,\widetilde{\rho}) > \epsilon_{robustness}$; that is, the deviation of the attacked path from the original path exceeds a given tolerance $\epsilon_{robustness}$.

\subsection{Definition of Resilience} \label{def_res}
For resilience, we take an ecological view widely seen in longitudinal population\cite{werner1993risk}, psychological\cite{dale2010community}, and biosystem\cite{folke2004regime} studies. 
\begin{framed}
Generally speaking, resilience indicates an innate capability to \emph{maintain} or \emph{recover} sufficient functionality in the face of challenging conditions \cite{black2008conceptual} against risk or uncertainty, while keeping a certain level of vitality and prosperity \cite{dale2010community}. 
\end{framed}

This definition of resilience does not consider the presence of risk as a parameter \cite{werner1993risk}, whereas the risks usually present themselves as uncertainty with heterogeneity in unpredictable directions including violence \cite{palmer2008theory}. 
The outcome of the resilience is usually evidenced by either: a recovery of the partial functionality, albeit possibly with a deviation from its designated features \cite{smith2008brief}; or a synthetisation of other functionalities with its adaptivity in congenital structure or inbred nature.
In the context of this paper, therefore resilience can be summarised as the system's ability to continue operation (even with reduced functionality) and recover in the face of adversity. In this light, robustness, \emph{inter alia}, is a feature of a resilient system. To avoid complicating discussions, we treat them separately.

In our definition, we take $dist^{e,e}(\rho,\widetilde{\rho})\leq \epsilon_{resilience}$ as the signal that the system has recovered to its designated functionality. Intuitively, $dist^{e,e}(\rho,\widetilde{\rho})\leq \epsilon_{resilience}$ means that the state estimation task has already returned back to normal -- within acceptable deviation $\epsilon_{resilience}$  on the time step $e \geq u$. 

Moreover, we take 
\begin{equation}
dist^{max} = \underset{t\in [l..u]}{\textbf{maximise}} ~  dist^{t,t}(\rho,\widetilde{\rho})
\end{equation} to denote the deviation of a path $\widetilde{\rho}$ from the normal path $\rho$. Intuitively,  it considers the maximum distance between two states: one on the original path and the other on the attacked path, at some time step $t$. The notation $max$ on $dist^{max}$ denotes the time step corresponding to the \emph{maximal} value. 

Then, the general idea of defining resilience for LE-SESs is that we measure the maximum deviation at some step $t\in [l,u]$ and want to know if the whole system can respond to the false information, gradually adjust itself, and eventually recover. Formally, taking $e \geq u$, we have the following formal definition of resilience:  

\begin{equation}
  \begin{array}{rl}
        \underset{\widetilde{\rho}}{\textbf{minimise}} & dist^{max}  \\

        \textbf{subject to} &  dist^{e,e}(\rho,\widetilde{\rho})  > \epsilon_{resilience} 
  \end{array}
  \label{equ:resiilencegeneral}
\end{equation}

Intuitively, the general optimisation objective is to minimize the maximum deviation such that the system cannot recover at the end of the path. In other words, the time $e$ represents the end of the path, where the state estimation functionality should have recovered to a certain level -- subject to the loss $\epsilon_{resilience}$.

We remark that, for resilience, the definition of ``recovery" can be varied. While in Equation (\ref{equ:resiilencegeneral}) we use $dist^{e,e}(\rho,\widetilde{\rho})  \leq \epsilon_{resilience}$ to denote the success of a recovery, there can be other definitions, for example, asking for a return to some path that does not necessarily have be the original one, so long as it is acceptable.

\subsection{Parameter Uncertainty}
The uncertainty in the \POsquare-LTS verification problem arises from three factors. First, different adversarial algorithms yield varying levels of accuracy when calculating payoff $\varphi$. We have selected the most popular and powerful adversarial algorithms, DeepFool \cite{moosavi2016deepfool} and DeepConcolic \cite{sun2018concolic}, to ensure an accurate measurement for the payoff. Second, the $L_p$ norm is employed to measure the deviation between the original path and the attacked path. Different values of $p$ can influence optimization results. We have chosen $p=2$, which represents the Euclidean distance, as it is widely accepted and utilized. Lastly, the definitions of tolerance, $\epsilon_{robustness}$ and $\epsilon_{resilience}$, have an impact on the verification results. These parameters can be determined empirically. For instance, one could use reference data gathered from practical scenarios to ascertain the optimal values for $\epsilon_{robustness}$ and $\epsilon_{resilience}$ that satisfy users' both functionality and safety requirements.

\subsection{Computational Complexity}
\label{sec:complexity}

We study the complexity of the \POsquare-LTS verification problem and show that the problem is NP-complete for both robustness and resilience. Concretely, for the soundness, an adversary can take a non-deterministic algorithm to choose the states $\widetilde{\rho}_k$ for a linear number of steps, and check whether the constraints satisfy in linear time. Therefore, the problem is in NP. For the resilience verification, it involves a min-max optimization problem, the solution of which has been proven to be NP-hard in \cite{daskalakis2021complexity}. To demonstrate the NP-hardness of robustness verification, we reduce the knapsack problem – a well-known NP-complete problem to the constrained optimization problem in Equation (\ref{equ:robustnessgeneral}).

We consider the 0-1 minimisation knapsack problem, which can be converted into the general knapsack problem form \cite{morales2020analysis}. The 0-1 minimisation knapsack problem restricts the number $c_i$ of copies of each kind of item to zero or one. Given a set of $n$ items numbered from 1 to $n$, each with a capacity $g_i$ and a cost $v_i$, and a minimum demand $W$, the objective is to compute 
\begin{equation}
    \begin{aligned}
        \underset{}{\textbf{minimise}}~~ & \sum_{i=1}^n v_ic_i\\
        \textbf{subject to}~~ &  \sum_{i=1}^n g_ic_i > W \\ 
        & \forall i\in [1..n]: c_i \in \{0,1\}
        \label{equ:knapsack}
    \end{aligned}
\end{equation}
where $c_i$ represents the number of the item $i$ to be included in the knapsack. Informally, this can be understood as the problem of buying items. The natural question is to minimise the sum of the cost of the items to buy while the overall capacity satisfies a minimum threshold
demand $W$. 

We can construct a \POsquare-LTS  $M=(Q,\rho^1_0,\KF,L,\payoff,\partialorder)$, where $Q=\{\rho^1_0\}\cup\bigcup_{i=1}^n\{\rho^0_{i},\rho^1_{i}\}$. Intuitively, for every item $i \geq 1$, we have two states representing whether or not the item is selected, respectively. Since item 0 is always selected, we denote it as $\rho^1_0$. For the transition relation $\KF$, we have that $\KF = \bigcup_{i=1}^n\{(\rho^1_{i-1},\rho^{c_i}_{i})|c_i\in \{0,1\}\}$, which connects each state of item $i-1$ to the states of the next item $i$. The payoff function $\payoff$ is defined as $\payoff(\rho^1_{i-1},\rho^0_{i})=0$ and $\payoff(\rho^1_{i-1}, \rho^1_{i})=v_i$, for all $(\rho^1_{i-1},\rho^{c_i}_{i})\in \KF$ and $c_i\in \{0,1\}$, representing that it will take $v_i$ payoff to take the transition $(\rho^1_{i-1},\rho^1_{i})$ in order to add the item $i$ into the knapsack, and take $0$ payoff to take the other transition $(\rho^1_{i-1},\rho^0_{i})$ in order to not add the item $i$. The partial order relation $\beta$ can be defined as having transition $(\rho^1_{i-1},\rho^0_{i}) \prec (\rho^1_{i-1},\rho^1_{i})$, for all $(\rho^1_{i-1}, \rho^{c_i}_{i})\in \KF$ and $c_i\in \{0,1\}$. 

For the specification, we have the following robustness-related optimisation objective. 
\begin{equation}
    \begin{array}{rl}
        \underset{\widetilde{\rho}}{\textbf{minimize}} & \varphi^{0,n}(\widetilde{\rho}) \\
        \textbf{subject to} & dist^{0,n}(\rho,\widetilde{\rho}) > W \\
    \end{array}
    \label{equ:knapsackoptimisation}
\end{equation}
such that 
\begin{equation}\label{eq:knapsackdistance}
dist(\rho_k,\widetilde{\rho}_k)=  \left\{
        \begin{array}{ll}
            0 & \text{ if } \widetilde{\rho}_{k} = \rho^0_{k} \\
            g_k & \text{ if } \widetilde{\rho}_{k} = \rho^1_{k}
        \end{array}
    \right.
\end{equation} 
Recall that,  $\varphi^{0,n}(\widetilde{\rho})$ is defined in Equations (\ref{equ:varphi}) and (\ref{eq:combined-payoff}), and $dist^{0,n}(\rho,\widetilde{\rho})$ is defined in Equation (\ref{eq:monitor-uncertainty}). As a result, the robustness of the model $M$ and the above robustness property is equivalent to the existence of a solution to the 0-1 Knapsack problem. This implies that the robustness problem on \POsquare-LTSs is NP-complete. 

\section{Automated Verification Algorithm}\label{sec:algorithms}

An attack on the LE-SESs, as explained in Section~\ref{percep}, adds perturbations to the input images in order to fool a neural network, which is part of the perception unit, into making wrong detections. On one hand, these wrong detections will be passed on to the perception unit, which in turn affects the Bayes filter and leads to wrong state estimation; the LE-SES can be vulnerable to such attack. On the other hand, the LE-SESs may have internal or external mechanisms to tolerate such attack, and therefore perform well with respect to properties such as robustness or resilience. It is important to have a formal, principled analysis to understand how good a LE-SES is with respect to the properties and whether a designed mechanism is helpful in improving its performance.

We have introduced in Section~\ref{opti_prob} how to reduce an LE-SES into a \POsquare-LTS $M$ and formally express a property -- either robustness or resilience -- with a constrained optimisation objective $\phi$ based on a path $\rho$ in Section \ref{sec:properties}. Thanks to this formalism, the verification of robustness and resilience can be outlined using the same algorithm. Now, given a model $M$, an optimisation objective $\phi$, and a pre-specified threshold $\theta$, we aim to develop an automated verification algorithm to check whether the model $M$ is robust or resilient on the path $\rho$; or formally, $sol_{opt}(M,\phi)>\theta$, where $sol_{opt}(M,\phi)$ denotes the optimal value obtained from the constrained optimisation problem over $M$ and $\phi$. 

The general idea of our verification algorithm is as follows. It first enumerates all possible paths of $M$ obtainable by attacking the given path $\rho$ (Algorithm~\ref{alg:exhausitive_search}), and then determines the optimal solution $sol_{opt}(M,\phi)$ among the paths (Algorithm~\ref{alg:robustness_resilience}). Finally, the satisfiability of the property is determined by comparing $sol_{opt}(M,\phi)$ and $\theta$. 

\subsection{Exhaustive Search for All Possible Paths}

The first step of the algorithm involves exhaustively enumerating all possible attacked paths on the \POsquare-LTS denoted as $M$ with respect to $\rho$. It is not difficult to observe that these paths form a tree structure, originating from the \POsquare-LTS $M$, as illustrated in Figure~\ref{fig:tree_diag}. Since a final deviation is not available until the end of a simulation, the tree must be fully expanded from the root to the leaf, and all the paths need to be explored. 

 \begin{algorithm}[htbp]
 \caption{Exhaustive Search based on BFS}
 \label{alg:exhausitive_search}
 \begin{algorithmic}[1]
 \renewcommand{\algorithmicrequire}{\textbf{Input:}}
 \renewcommand{\algorithmicensure}{\textbf{Output:}}
 \REQUIRE  LTS model $M$, $n$, $l$, $u$
 \ENSURE path set $P$, payoff set $\varphi^{l,u}$
 \STATE run original path $\rho$ from $k = 0$ to $k = n$
 \STATE set $\rho_{l-1}$ as root node
 \FOR {$k$ from $l$$-$$1$ to $u$} 
 \FOR {each node $\widetilde{\rho}_{k}$ in $leaf(\rho_{l-1})$}
 \STATE find potential observations $Z \leftarrow neighbours(\widetilde{\rho}_{k})$
 \FOR {each observed location $\matr{z}$ in $Z$}
 \STATE $\widetilde{\rho}_{k+1} \leftarrow kf(\widetilde{\rho}_k,\matr{z})$
 \STATE calculate the attack payoff $\varphi(\widetilde{\rho}_{k+1}, \widetilde{\rho}_k)$
 \STATE $\widetilde{\rho}_{k} = parent(\widetilde{\rho}_{k+1})$
 \ENDFOR
 \ENDFOR
 \ENDFOR
 \STATE $P \leftarrow path(\rho_{l-1})$
 \STATE run path $\widetilde{\rho}$ in set $P$ to $k = n$
 \STATE calculate the combined payoff for each path $\widetilde{\rho}$  $\varphi^{l,u} (\widetilde{\rho}) = \sum_{k=l}^u \varphi(\widetilde{\rho}_{k-1},\widetilde{\rho}_{k})/(u-l) $
 \RETURN $P$, $\varphi^{l,u}$
 \end{algorithmic} 
\end{algorithm}

Suppose the tree has $V$ nodes and a depth of $n$. Clearly, if all paths are calculated recursively from the top to the end, the time complexity of this procedure is exponential in terms of the number of depths, denoted as $\mathcal{O}(2^n)$. This is consistent with our complexity result, as presented in Section~\ref{sec:complexity}. To avoid the re-computation of tree nodes, we employ the breadth-first search (BFS) method \cite{kozen1992depth} to enumerate the paths. By storing the results of sub-problems, the time complexity can be reduced to a linear complexity of $\mathcal{O}(V)$.

The details are presented in Algorithm~\ref{alg:exhausitive_search}. We need several operation functions on the tree, including $leaf$ (which returns all leaf nodes of the root node), $parent$ (which associates a node to its parent node), and $path$ (which returns all tree paths from the given root node to the leaf nodes).
 
Lines 2-12 in Algorithm~\ref{alg:exhausitive_search} present the procedure of constructing the tree diagram. First, we set the root node $\rho_{l-1}$ (Line 2), that is, we will attack the system from the $l-1$ state of the original path $\rho$ and enumerate all possible adversarial paths. At each step $k$, function $neighbours$ will list all observations near the predicted location (Line 5). Then, each observation is incorporated with current state $\matr{\rho}_k$, which is stored in memory, for the calculation of the next state $\matr{\rho}_{k+1}$ (Line 7). If no observation is available or $\matr{z} = \emptyset$ , the KF can still run normally, skipping the update phase. To enable each transition $(\matr{\rho}_k,\matr{\rho}_{k+1})$, the partial order relation is followed when attacking the system and recording the payoff $\varphi$ (Line 8). Then, the potential $\matr{\rho}_{k+1}$ is accepted and added as the child node of $\matr{\rho}_k$. Once the tree is constructed, we continue simulating the paths to the end of time, $k=n$, (Lines 13-14). Finally, all the paths in set $P$ are output along with the attack payoff $\varphi^{l,u}$ (Lines 15-16). 

\subsection{Computing an Optimal Solution to the Constrained Optimisation Problem}\label{sec:solveoptimisation}

After enumerating all possible paths in $P$, we can compute optimal solutions to the constrained optimisation problems as in Equation (\ref{equ:robustnessgeneral}) and (\ref{equ:resiilencegeneral}). We let $obj$ be the objective function to minimize, and $con$ be the constraints to follow. For robustness,
we have $obj = \varphi^{l,u}$ and $con = dist^{0,e}$, and for resilience, we have 
$obj = dist^{max}$ and $con = dist^{e,e}$.

\begin{algorithm}
 \caption{Computation of Optimal Solution and A Representative Path}
 \label{alg:robustness_resilience}
 \begin{algorithmic}[1]
 \renewcommand{\algorithmicrequire}{\textbf{Input:}}
 \renewcommand{\algorithmicensure}{\textbf{Output:}}
 \REQUIRE  path set $P$, $obj$, $con$, $\epsilon$
 \ENSURE representative path $\widetilde{\rho}^*$, $obj$ value $\theta^*$ of $\rho^*$, and optimal value $sol_{opt}(M,\phi)$
 \STATE find the original path $\rho$ in set $P$
 \STATE $sol_{opt}(M,\phi) \leftarrow 0$, $k \leftarrow 0$, $P^+ \leftarrow \emptyset$, $P^- \leftarrow \emptyset$
 \FOR {$\widetilde{\rho}$ in set $P$} 
 \IF {$con(\rho,\widetilde{\rho}) > \epsilon$}
 \STATE $k \leftarrow k+1$
 \STATE $P^+ \leftarrow P^+ \cap \{\widetilde{\rho}\}$
 \IF {$k = 1$ or $obj(\rho,\widetilde{\rho}) < sol_{opt}(M,\phi)$}
 \STATE $sol_{opt}(M,\phi) \leftarrow obj(\rho,\widetilde{\rho})$
 \ENDIF
 \ELSE 
 \STATE $P^- \leftarrow P^- \cap \{\widetilde{\rho}\}$
 \ENDIF
 \ENDFOR
 \STATE $\rho^* \leftarrow \arg\max_{\widetilde{\rho} \in P^{-}, obj < sol_{opt}(M,\phi)}  obj(\rho,\widetilde{\rho}) $
 \STATE $\theta^* \leftarrow obj(\rho,\rho^*) $
 \RETURN $\rho^*$, $\theta^*$, $sol_{opt}(M,\phi)$ 
 \end{algorithmic} 
\end{algorithm}

Note that, our definitions in Equations (\ref{equ:robustnessgeneral}) and (\ref{equ:resiilencegeneral}) are set to work with cases that do not satisfy the properties, i.e., paths that are not robust or resilient, and identify the optimal one from them. Therefore, a path satisfying the constraints suggests that it does not satisfy the property. We split the set $P$ of paths into two subsets, $P^+$ and $P^-$. Intuitively, $P^+$ includes those paths satisfying the constraints, i.e., fail to perform well with respect to the  property, and $P^-$ includes those paths that do not satisfy the constraints, i.e., perform well with respect to the property. For robustness, $P^+$ includes paths satisfying  $dist^{0,e}(\rho,\widetilde{\rho}) > \epsilon$ and $P^-$ satisfying $dist^{0,e}(\rho,\widetilde{\rho}) \leq \epsilon$. For resilience, $P^+$ includes paths satisfying  $dist^{e,e}(\rho,\widetilde{\rho}) > \epsilon$ and $P^-$ satisfying $dist^{e,e}(\rho,\widetilde{\rho}) \leq  \epsilon$. 

In addition to the optimal solutions that, according to the optimisation objectives, are some of the paths in $P^+$, it is useful to identify certain paths in $P^-$ that are robust or resilient. Let $sol_{opt}(M,\phi)$ be the optimal $obj$ value, from the optimal solution. We can sort the paths in $P^-$ according to their $obj$ value, and let \textbf{representative path} $\rho^{*}$ be the path whose $obj$ value is the greatest among those smaller than $sol_{opt}(M,\phi)$. Intuitively, $\rho^{*}$ represents the path that is closest to the optimal solution of the optimisation problem but satisfies the corresponding robust/resilient property. This path is representative because it serves as the worst case scenario for us to exercise the system's robust property and resilient property respectively. Moreover, we let $\theta^*$ be the $obj$ value of robustness/resilience of the path $\rho^*$, called \textbf{representative value} in the paper.

The algorithm for the computation of the optimal solution and a representative path can be found in Algorithm \ref{alg:robustness_resilience}. Lines 1 to 9 give the process to solve Equation (\ref{equ:robustnessgeneral}) or (\ref{equ:resiilencegeneral}) for the optimal value $sol_{opt}(M,\phi)$. The remaining Lines calculate the representative value $\theta^*$ and a representative path $\rho^*$.

For each adversarial path in $P$, it is added into either $P^+$ (Line 6) or $P^-$ (Line 11). The minimum objective function is then found by comparing the adversarial tracks in set $P^+$ (Lines 4-9). We need to find the representative path in set $P^-$, which has the $obj$ value smaller than $sol_{opt}(M,\phi)$ but lager than any other path in $P^-$ (Line 14). Its corresponding representative value is  computed  in Line 15. 

\section{Case Study: A Real-World WAMI Dynamic Tracking System} \label{sec:WAMI_intro}

In this part, we present a brief introduction, followed by the technical details, to the real-world WAMI dynamic tracking system that will be used as major case study. The tracking system requires continuous imagery input from e.g., airborne high-resolution cameras. In the case study, the input is a video, which consists of a finite sequence of WAMI images. Each image contains a number of vehicles. The essential processing chain of the WAMI tracking system is as follows.

\begin{enumerate}[leftmargin=*]
    \item Align a set of previous frames with the incoming frame.
    \item Construct the background model 
    of incoming frames using the median frame.
    \item Extract moving objects using background subtraction.
    \item Determine if the moving objects are  vehicles by using a Binary CNN.
    \item For complex cases, predict the locations of moving objects/vehicles using a regression CNN.
    \item Track one of the vehicles using a Kalman filter. 
\end{enumerate}

WAMI tracking uses \textbf{Gated nearest neighbour (Gnn)} to choose the new observation $\matr{z}_k$: from the set $Z_k$,  the one closest to the predicted measurement $\matr{H}_{k} \cdot \hat{\matr{s}}_{k}$ is chosen, i.e., 
\begin{align}\label{equ:WAMIobservation}
    &\matr{z}_k = \arg\min_{\matr{z}\in Z_k} ||\matr{z} - \matr{H}_{k} \cdot \hat{\matr{s}}_k||_p \\
    &s.t. \quad ||\matr{z} - \matr{H}_{k} \cdot \hat{\matr{s}}_k||_p \leq  \tau_k
\end{align}
where $||\cdot||_p$ is $L_p$-norm distance ($p=2$, i.e., Euclidean distance is used in this paper), and $\tau_k$ is the gate value, representing the maximum uncertainty in which the system is able to work. 

Specifically, the WAMI system has the following definitions of  $\matr{s}$ and  $\matr{P}$:
\begin{equation}
\begin{array}{lcl}
\matr{s} & = 
\begin{bmatrix}
\matr{l} \\ \matr{v}
\end{bmatrix}
~~~~~~
\matr{P} & = 
\begin{bmatrix}
\Sigma_{\matr{l}\matr{l}} & \Sigma_{\matr{l}\matr{v}} \\
\Sigma_{\matr{v}\matr{l}} & \Sigma_{\matr{v}\matr{v}} 
\end{bmatrix}
\end{array}
\end{equation}
where $\matr{s}$ denotes the mean values of two Gaussian stochastic variables, $\matr{l}$ representing the location which is measurable from the input videos, and $\matr{v}$ representing the velocity which cannot be measured directly, respectively.

In the measurement space, the elements in $\matr{l}$ are not correlated, which makes it possible to simplify the Bayesian uncertainty metric, $\tau$, that is the trace of the covariance matrix:

\begin{equation} \label{uncertainty}
\tau = tr(\Sigma_{\matr{l}\matr{l}})
\end{equation}

Therefore, $\tau$ is partially related to the search range in which observations can be accepted. Normally, $\tau$ will gradually shrink before being bounded --  the convergence property of KF.

\subsection{Wide-Area Motion Imagery Input}
\label{sec:wami-detector-input}

The input to the tracking system is a video comprising a finite sequence of images, with each image containing a number of vehicles. Following the approach in \cite{ZM2019}, we employ the WPAFB 2009 dataset~\cite{7158947}. These images were captured by a camera system equipped with six optical sensors, which were pre-stitched to cover a broad area approximately $35 km^{2}$ in size. The dataset operates at a frame rate of 1.25Hz and consists of 1025 frames. This translates to about 13 minutes of footage, partitioned into training ($512$ frames) and testing ($513$ frames) videos. Importantly, every vehicle and its trajectory within this dataset has been manually annotated. Multiple video resolutions are available in the dataset. For our experiment, we opted for the $12,000 \times 10,000$ resolution images, where vehicle sizes are less than $10 \times 10$ pixels. We use the notation $\videoframe_i$ to denote the $i$-th frame, and $\videoframe_i(x,y)$ represents the pixel located at the intersection of the $x$-th column and $y$-th row of $\videoframe_i$.

In the subsequent sections, we elucidate the workings of the tracking system that takes video as its input. Broadly, this process is bifurcated into two stages: detection and tracking. From Section~\ref{sec:registration} to Section~\ref{sec:framework}, we detail the detection steps, focusing on how vehicles are detected using CNN-based perception units. The tracking procedure is then elaborated upon in Section~\ref{sec:wami-tracker}. 

\subsection{Background Construction}\label{sec:registration}

Vehicle detection in WAMI video presents challenges due to the limited visibility of vehicle appearances and the presence of frequent pixel noise. As discussed in \cite{SommerTSB16,LaLondeZS18}, appearance-based object detectors can result in a high number of false alarms. Because of this, this paper focuses solely on the detection of moving objects for tracking purposes.

Constructing a background is an essential step for extracting pixel changes from an input image. The background is established based on the current environment using a series of previous frames captured by the moving camera system. This is accomplished through the following steps:

\paragraph{Image registration} This process compensates for camera motion by aligning all previous frames to the current frame. The essential step is estimating a transformation matrix, $h_{k}^{k-t}$, that transforms frame $\videoframe_{k-t}$ to frame $\videoframe_k$ using a designated transformation function. We employ projective transformation (or homography) as this transformation function. It has been extensively used in multi-perspective geometry, a domain where WAMI camera systems are already prevalent.

The estimation of $h_{k}^{k-t}$ is generated by applying  
feature-based approaches. First of all, feature points from images at frame $\videoframe_{k-t}$ and $\videoframe_k$, respectively, are extracted by feature detectors (e.g., Harris corner or SIFT-like~\cite{SIFT} approaches). Second, feature descriptors, such as SURF~\cite{bay2006surf} and ORB~\cite{rublee2011orb}, are computed for all detected feature points. Finally, pairs of corresponding feature points between two images can be identified and the matrix $h_{k}^{k-t}$ can be estimated by using RANSAC~\cite{fischler1981random}, which is robust against outliers.

\paragraph{Background Modeling} We generate the background, $\matr{I}_{k}^{bg}$, for each time $k$, by computing the median image of the $L$ previously-aligned frames, i.e., 
\begin{equation}
\matr{I}_{k}^{bg}(x,y) = \frac{1}{L}\sum_{i=1}^L \videoframe_{k-i}(x,y)
\end{equation}
In our experiments, we take either  $L=4$ or $L=5$.

Note that, to align the $L$ previous frames to the newly received frame, only one image registration process is performed. After obtaining the matrices $h_{k-1}^{k-2}, h_{k-1}^{k-3}, ...$ by processing previous frames, we perform image registration once to get $h_{k}^{k-1}$, and then let 
\begin{equation}\label{equ:updatetemplate}
h_{k}^{k-2}=h_{k}^{k-1} \times h_{k-1}^{k-2},~  h_{k}^{k-3}=h_{k}^{k-1} \times h_{k-1}^{k-3}.
\end{equation}

\paragraph{Extraction of Potential Moving Objects} By comparing the difference between $\matr{I}_k^{bg}$ and the current frame $\alpha_k$, we can extract a set $Q_{bc}$ of potential moving objects  by first computing the following set of pixels
\begin{equation}
    P_{bc} = \{ (x,y)~|~ |\matr{I}_k^{bg}(x,y) - \alpha_k(x,y)| > \delta_{bc}, (x,y) \in \Gamma \}
\end{equation}
and then applying image morphology operation on
$P_{bc}$, where $\Gamma$ is the set of pixels and  $\delta_{bc}$ is a threshold value to determine which pixels should be considered. 

\subsection{CNN for Detection Refinement}\label{sec:refinement}

After obtaining $P_{bc}$, we develop a CNN, $\network_{dr}$, to detect vehicles. We highlight a few design decisions. 
The major causes of false alarms generated by the background subtraction are: poor image registration, light changes and the parallax effect in high objects (e.g., buildings and trees). We emphasise that the objects of interest (e.g., vehicles) mostly, but not exclusively, appear on roads. Moreover, we perceive that a moving object generates a temporal pattern (e.g., a track) that can be exploited to discern whether or not a detection is an object of interest. Thus, in addition to the shape of the vehicle in the current frame, we assert that the historical context of the same place can help to distinguish the objects of interest and false alarms. 

By the above observations, we create a binary classification CNN $\network_{dr}:\real^{21\times 21\times (m+1)}\xrightarrow{} \{0,1\}$ to predict whether a $21 \times 21$ pixels window contains a moving object given aligned image patches generated from 
the previous $m$ frames. The $21 \times 21$ pixels window is identified by considering the image patches from the set $Q_{bc}$. We suggest $m=3$ in this paper, as it is the maximum time for a vehicle to cross the window. The input to the CNN is a $21 \times 21 \times (m+1)$ matrix and the convolutional layers are identical to traditional 2D CNNs, except that the three colour channels are substituted with $m+1$ grey-level frames.

Essentially, $\network_{dr}$ acts as a filter to remove, from $Q_{bc}$, objects that are unlikely to be vehicles. Let $Q_{dr}$ be the obtained set of moving objects. If the size of an image patch in $Q_{dr}$ is similar to a vehicle, we directly label it as a vehicle. On the other hand, if the size of the image patch in $Q_{dr}$ is larger than a vehicle, i.e., there may be multiple vehicles, we pass this image patch to the location prediction for further processing.

\subsection{CNN for Location Prediction}\label{sec:position}

We use a regression CNN  $\network_{lp}:\real^{45\times 45\times (m+1)}\xrightarrow{}\real^{15\times 15}$ to process image patches passed over from the detection refinement phase.
As in~\cite{LaLondeZS18}, a regression CNN can predict the locations of objects given spatial and temporal information. The input to $\network_{lp}$ is similar to the classification CNN $\network_{dr}$ described in Section~\ref{sec:refinement}, except that the size of the window is enlarged  to $45 \times 45$. The output of $\network_{lp}$ is a $225$-dimensional vector, equivalent to a down-sampled image ($15 \times 15$) for reducing computational cost. 

For each $15 \times 15$ image, we apply a filter to obtain those pixels whose values are greater than not only a threshold value $\delta_{lp}$ but also the values of its adjacent pixels. We then obtain another $15 \times 15$ image with a few bright pixels, each of which is labelled as a vehicle. Let $O$ be the set of moving objects updated from $Q_{dr}$ after applying location prediction. 

\subsection{Detection Framework}\label{sec:framework}

\begin{figure*}[!ht]
\centering
\includegraphics[width=\textwidth]{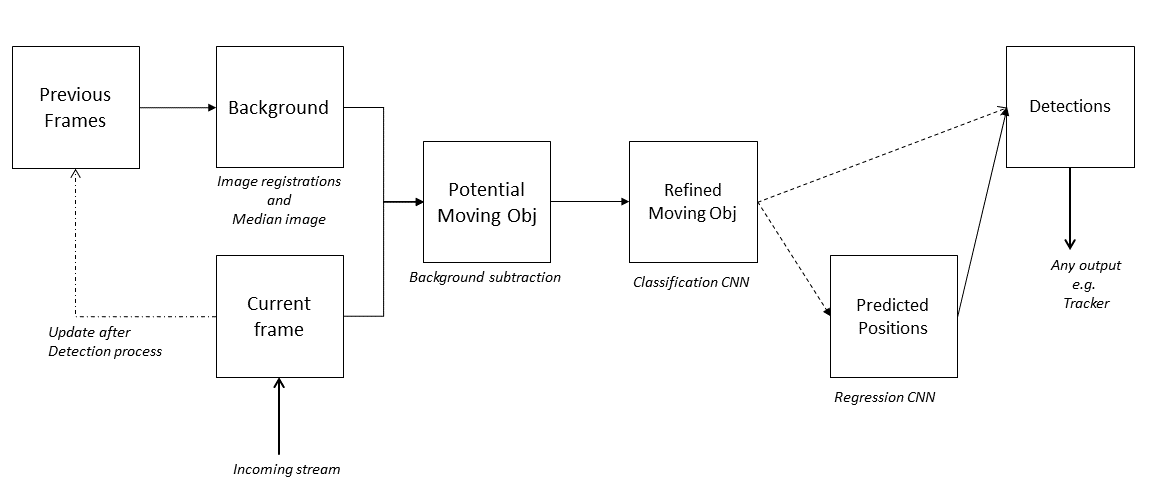}
\caption{The architecture of the vehicle detector. \label{fig:detector}}
\end{figure*}
The processing chain of the detector is shown in Figure~\ref{fig:detector}. At the beginning of the video, the detector takes the first $L$ frames to construct the background, thus the detections from frame $L+1$ can be generated. After the detection process finishes in each iteration, it is added to the template of previous frames. The updating process substitutes the oldest frame with the input frame. This is to ensure that the background always considers the latest scene, since the frame rate is usually low in WAMI videos such that parallax effects and light changes can be pronounced. As we wish to detect very small and unclear vehicles, we apply a small background subtraction threshold and a minimum blob size. This, therefore, leads to a huge number of potential blobs. The classification CNN is used to accept a number of blobs. As mentioned in Section~\ref{sec:refinement}, the CNN only predicts if the window contains a moving object or not. According to our experiments, the cases where multiple blobs belong to one vehicle and one blob includes multiple vehicles, occur frequently. Thus, we design two corresponding scenarios: the blob is very close to another blob(s); the size of the blob is larger than $20 \times 20$. If any blob follows either of the two scenarios, we do not consider the blob for output. The regression CNN (Section~\ref{sec:position}) is performed on these scenarios to predict the locations of the vehicles in the corresponding region, and a default blob will be given. If the blob does not follow any of the scenarios, this blob will be output directly as a detection. Finally, the detected vehicles include the output of both sets.

\subsection{Object Tracking}
\label{sec:wami-tracker}

\subsubsection{Problem Statement}

We consider a single target tracker (i.e. Kalman filter) to track a vehicle given all the detection points over time in the field of view. The track is initialised by manually giving a starting point and a zero initial velocity, such that the state vector is defined as $s_{t}=[x_t, y_t, 0, 0]^T$ where $[x_t, y_t]$ is the coordinate of the starting point. We define the initial covariance of the target, $P=diag \left[30, 30, 20, 20 \right]^2$, which is the initial uncertainty of the target's state\footnote{With this configuration, the starting point does not need to be the precise position of a vehicle; the tracker will identify a nearby target to track. However, by defining a specific velocity and reducing the uncertainty in $\matr{P}$, it becomes possible to track a specific target.}.

A near-constant velocity model is applied as the dynamic model in the Kalman filter. This model is defined by concretizing Equation~(\ref{equ:process_model1}):
\begin{equation}
\hat{\matr{s}}_{k} = \matr{F} \cdot \matr{s}_{k-1} + \omega_k~~~\text{ s.t. }    \matr{F} =
\left[ 
\begin{array}{cccc}
    \mathbb{I}_{2\times2} & \mathbb{I}_{2\times2} \\
    \mathbb{O}_{2\times2} & \mathbb{I}_{2\times2} \\
\end{array}
\right]
\label{eq:dynamics-model}
\end{equation}
\noindent where $\mathbb{I}$ is a identity matrix, $\mathbb{O}$ is a zero matrix, and $\matr{s}_{k-1},\hat{\matr{s}}_{k-1}$ and $\omega_k$ are as defined in Section~\ref{sec:preliimnaries}, such that the covariance matrix $\matr{Q}$ of the process noise is defined as follows:
\begin{equation}
\matr{Q} = \sigma_{q}^{2} \cdot 
    \begin{bmatrix}
    \frac{1}{3} dt^{3} \cdot \mathbb{I}_{2\times2} & \frac{1}{2} dt^{2}  \cdot \mathbb{I}_{2\times2} \\
    \\
    \frac{1}{2} dt^{2} \cdot \mathbb{I}_{2\times2} &  \mathbb{I}_{2\times2}
    \end{bmatrix}
\label{eq:processnoiseQ}
\end{equation}
\noindent where $dt$ is the time interval between two frames and $\sigma_q$ is a configurable constant. $\sigma_q=3$ is suggested for the aforementioned WAMI video.

Next, we define the measurement model by concretising Equation~(\ref{equ: measuremnet_model}):
\begin{equation}
\matr{z}_k = \matr{H} \cdot  \matr{s}_{k} + \matr{v}_k \text{\quad s.t. \quad}\matr{H} = \left[ 
\begin{array}{cccc}
    1 & 0 & 0 & 0 \\
    0 & 1 & 0 & 0 \\
\end{array}
\right]
\label{eq:measurementmodel}
\end{equation}
\noindent where $\matr{z}_k$ is the measurement, representing the position of the tracked vehicle, and $\matr{s}_k$ and $\matr{v}_k$ are defined in Section~\ref{sec:preliimnaries}. The covariance matrix, $\matr{R}$, is defined as $\matr{R} = \sigma_{r}^{2} \cdot \mathbb{I}_{2\times2}$, where we suggest $\sigma_{r}=5$ for the WAMI video.

Since the camera system is moving, the position should be compensated for such motion using the identical transformation function for image registration. However, we ignore the influence to the velocity as it is relatively small, but consider integrating this into the process noise.

\subsubsection{Measurement Association}

During the update step of the Kalman filter, the residual measurement should be calculated by subtracting the measurement ($\matr{z}_k$) from the predicted state ($\hat{\matr{s}}_{k}$). In the tracking system, a Gnn is used to obtain the measurement from a set of detections. K-nearest neighbour is first applied to find the nearest detection, $\hat{\matr{z}}_{k}$, of the predicted measurement, $\matr{H} \cdot  \hat{\matr{s}}_{k}$. Then the Mahalanobis distance between $\hat{ \matr{z}_k}$ and $\matr{H} \cdot  \hat{\matr{s}}_{k}$ is calculated as follows:
\begin{equation}
    D_{k} = \sqrt{ (\hat{\matr{z}}_k-\matr{H} \cdot  \hat{\matr{s}}_{k})^T \cdot \hat{\matr{S}_{k}}^{-1} \cdot (\hat{\matr{z}}_k-\matr{H} \cdot  \hat{\matr{s}}_{k})}
    \label{eq:Mdist}
\end{equation}
where $\hat{\matr{P}_{k}}$ is the innovation covariance, which is defined within the Kalman filter.

A potential measurement is adopted if $D_{k} \leq g$ with $g = 2$ in our experiment. If there is no available measurement, the update step will not be performed and the state uncertainty accumulates. It can be noticed that a large covariance leads to a large search window. Because the search window can be unreasonably large, we halt the tracking process when the trace of the covariance matrix exceeds a pre-determined value.

\section{Improvements to WAMI Tracking System}\label{sec:design2}

In this section, we introduce two techniques to improve the robustness and resilience of the LE-SESs. One of the techniques uses a runtime monitor to track a convergence property, expressed with the covariance matrix $\matr{P}_{k}$; and the other considers components to track multiple objects around the primary target to enhance fault tolerance in the state estimation. 

\subsection{Runtime Monitor for Bayesian Uncertainty}\label{sec:uncertaintymonitor}

Generally speaking, a KF system includes two phases: prediction (Equation (\ref{eq:kf_predict})) and update (Equation (\ref{eq:kf_update})). Theoretically, a KF system can converge \cite{Kalman1960} with optimal parameters: $\matr{F}$, $\matr{H}$, $\matr{Q}$, and $\matr{R}$, that well describe the problem. 
In this paper, we assume that the KF system has been well designed to ensure the convergence. Empirically, this has been proven possible in many practical systems. We are interested in another characteristic of the KF: when no observation is available, and therefore the update phase isn't performed, the uncertainty, $\hat{\matr{P}}_{k}$, grows relative to $\matr{P}_{k-1}$. In such instances, the predicted covariance $\hat{\matr{P}}_{k}$ is treated as the updated covariance $\matr{P}_{k}$ for that time-step.

In the WAMI tracking system, if a track lacks associated observations (e.g., due to mis-detections) for a certain period, the magnitude of the uncertainty metric $\tau$ accumulates and may eventually 'explode'. Consequently, the search range for observations expands dramatically. This scenario can be leveraged to design a monitor for measuring the attack, and it should be factored in when analyzing the system's robustness and resilience.

The monitor for the Bayesian uncertainty can be designed as follows: if $\tau$ increases, an alarm is activated to signal a potential attack. From the attacker's viewpoint, to evade this alert, any successful attack should aim to obscure the rise in $\tau$. To discern when this increment might manifest in the WAMI tracking system, we revisit the discussion in Section \ref{sec:WAMI_intro}. Here, a track is associated with the nearest observation $\matr{z}$, based on the Mahalanobis distance, within a predefined threshold for each time step. By targeting all the observations in $Z_k$, one can craft a scenario where no observations fall within the search range. This mirrors the previously described situation where the Bayesian uncertainty metric surges due to a missed update phase.

Formally, we introduce a parameter $\Gamma$, as defined in (\ref{eq:gamma}), which monitors the variations in the Kalman filter's covariance over time and takes into account the convergence process.
\begin{equation}
   \Gamma(\widetilde{\rho}_{k-1},\widetilde{\rho}_{k}) =  \left\{
        \begin{array}{ll}
            1 & \quad \tau(\widetilde{\rho}_{k}) \leq \tau(\widetilde{\rho}_{k-1}) \\
            0 & \quad \tau(\widetilde{\rho}_{k}) > \tau(\widetilde{\rho}_{k-1})
        \end{array}
    \right.
    \label{eq:gamma}
\end{equation}

\subsection{Joining Collaborative Components for Tracking}\label{multi_kf}

The previous WAMI tracking system experienced malfunctions primarily due to two causes: false alarms and mis-detections. It's important to note that, in this paper, we focus exclusively on tracking a single target. Utilizing a Gnn and a Kalman filter effectively handles false alarms in the majority of instances. However, mis-detections still present significant challenges. As mentioned in Section~\ref{sec:uncertaintymonitor}, mis-detections may cause the Bayesian uncertainty range to expand. Given the high number of detections typically found in WAMI videos, there's a risk that tracking could inadvertently shift to a different target. To address this challenge, we propose an approach that \textbf{\emph{utilizes joining collaborative components}}, which we refer to as joint Kalman filters (joint-KFs). In this approach, we simultaneously track several targets in close proximity to the main target using multiple Kalman filters. The detailed methodology is outlined as follows:

\begin{itemize}[leftmargin=*]
    \item Two kinds of Kalman filter tracks are maintained: one track for the primary target, $T^{p}$, and multiple tracks for the refining association, $T^{r}$.
    \item At each time-step other than the initialisation step, we have predicted tracks  $\{ \hat{T_{k}}^{p}, \hat{T_{k}}^{r} \}$, a set $Z_k$ of detections from current time-step, and a set $\tilde Z_{k-1}$ of unassociated detections from the previous time-step.
    \begin{enumerate}
        \item Calculate the likelihoods of all the pairs of detections $Z_k$ and tracks $\{ \hat{T_{k}}^{p}, \hat{T_{k}}^{r} \}$ using $\mathcal{N}\left( \hat{\matr{s}}_k, \matr{z}_k, \matr{S}_k \right)$ where the parameters can be found in Kalman filter.
        \item Sort the likelihoods from the largest to the smallest, and do a gated one-to-one data association in this order.
        \item Perform standard Kalman filter updates for all the tracks.
        \item For each detection in $Z_k$ that is not associated and is located close to the primary target, calculate the distance to each element in
        $Z_{k-1}$.
        \begin{itemize}
            \item If the distance is smaller than a predefined value, initialize a track and treat the distance as velocity then adding this track into $T_{k}^{r}$.
            \item Otherwise, store this detection in $\tilde Z_{k}$
        \end{itemize}
        \item Maintain all the tracks for refining association, $T_{k}^{r}$: if a track is now far away from the primary target, remove it from the set.
    \end{enumerate}
\end{itemize}

By applying this data association approach, if the primary target is mis-detected, the track will not be associated to a false detection, and even when this occurs for a few time-steps and the search range becomes reasonably large, this system can still remain resilient (i.e. can still function and recover quickly). 

\section{Experimental Evaluation}\label{sec:experiments}

We conduct an extensive set of experiments to show the effectiveness of proposed verification algorithm in the design of the WAMI tracking system. We believe that our approaches can be easily generalised to work with other autonomous systems using both Bayes filter(s) and neural networks.

\subsection{Research Questions}
Our evaluation experiments are guided by the following  research questions.
\begin{itemize}
\item[\textbf{RQ1}] What is the evidence of system level robustness and resilience for the WAMI tracking in Section \ref{sec:WAMI_intro}? 
\item[\textbf{RQ2}] What are the differences and similarity between robustness and resilience within the WAMI tracking?
\item[\textbf{RQ3}] Following \textbf{RQ1} and \textbf{RQ2}, can our verification approach be used to identify, and quantify the risk to the robustness and resilience of the WAMI tracking system? 
\item[\textbf{RQ4}] Are the improved design presented in Section~\ref{sec:design2} helpful in improving the system's robustness and resilience?
\end{itemize}

Their respective experiments and results are presented in Section~\ref{sec:expevidence}, \ref{sec:expcomparison},  \ref{sec:verificationexperiments} and \ref{sec:expimprovement}. 

\subsection{Experimental Setup}

We consider a number of original tracks with maximum length of 20 steps ($e = 19, k\in[0,19]$). An attack on the system is conducted between time steps $l$ and $u$, denoted as $Attack(l,u)$, with the following configurations: $l \in [4,12]$, and $(u-l) \in [1,4]$. The original track is coloured in \emph{green} in both the high-resolution images (Figure~\ref{robustness_example}--\ref{fig:multi_kf}) and the state space unfolding (Figure~\ref{fig:state_space}). The attacked track is coloured in \emph{red}. The white-colour arrows in the high-resolution images indicate the ground-truth directions of the vehicle. 

Moreover, in all experiments, for every attacked track, we record the following measures: the combined payoff $\varphi$, the cumulative deviation $dist^{0, e}$, the maximum deviation $dist^{max}$ and the final step deviation $dist^{e,e}$, such that $ m$ is the time step for maximum deviation and $e$ is the end of the track.

\subsection{Evidence of System's Robustness and Resilience}\label{sec:expevidence}

\begin{figure}[ht] 
  \begin{subfigure}[b]{0.5\linewidth}
    \centering
    \includegraphics[width=0.9\linewidth]{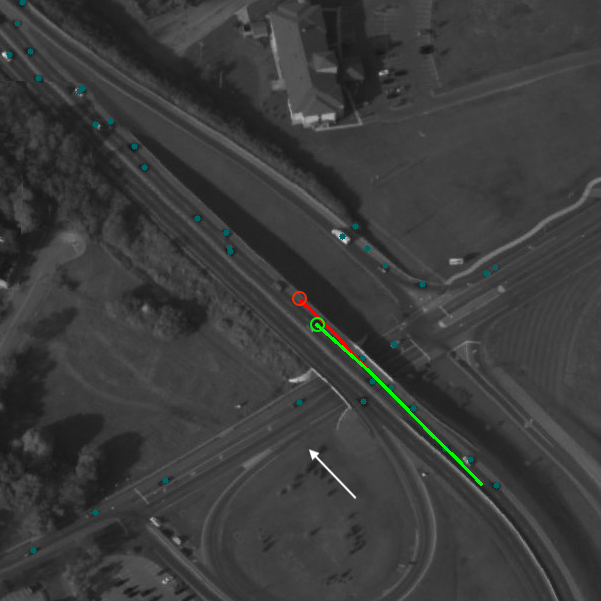} 
    \caption{$k = 6$, $\varphi = 10.47$} 
    \label{robustness_example:a} 
    \vspace{2ex}
  \end{subfigure}
  \begin{subfigure}[b]{0.5\linewidth}
    \centering
    \includegraphics[width=0.9\linewidth]{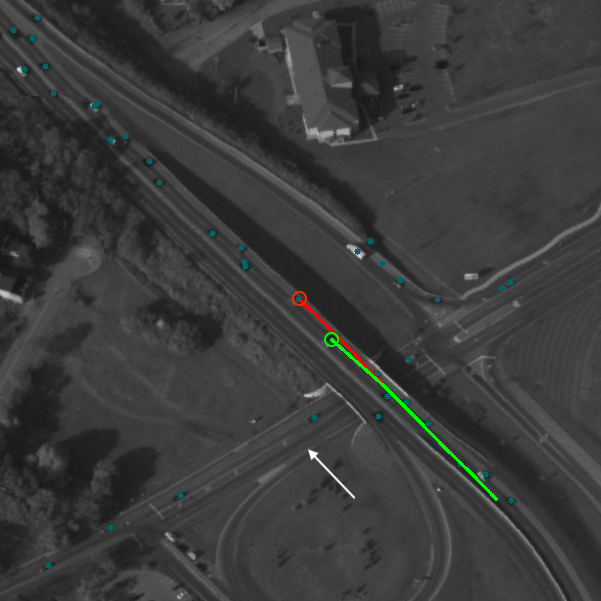} 
    \caption{$k = 6$, $\varphi = 20.58$}
    \label{robustness_example:b} 
    \vspace{2ex}
  \end{subfigure} 
  \begin{subfigure}[b]{0.5\linewidth}
    \centering
    \includegraphics[width=0.9\linewidth]{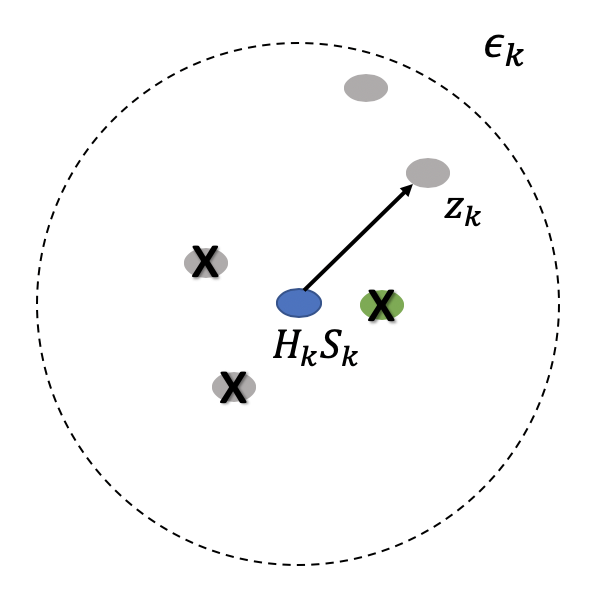} 
    \caption{KF's update process for (a)} 
    \label{robustness_explain:a} 
  \end{subfigure}
  \begin{subfigure}[b]{0.5\linewidth}
    \centering
    \includegraphics[width=0.9\linewidth]{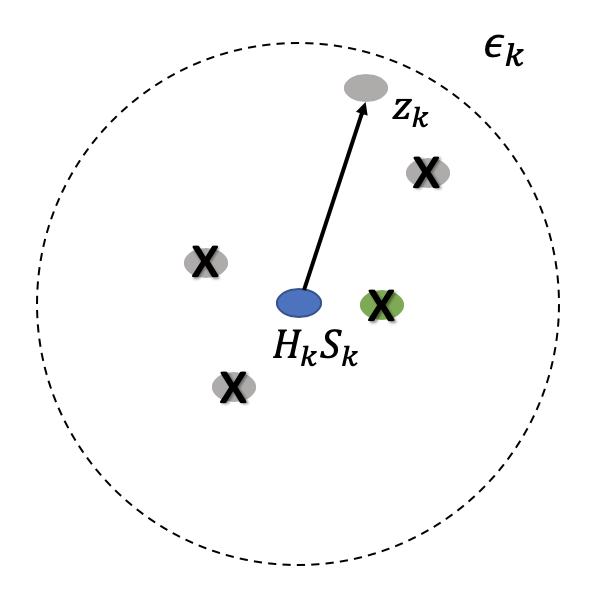} 
    \caption{KF's update process for (b)} 
    \label{robustness_explain:b} 
  \end{subfigure} 
  \caption{Illustration of WAMI system's robustness to the consecutive attack at $k = 5-6$. With increased attack payoff, the deviation is still bounded.}
  \label{robustness_example} 
\end{figure}

To demonstrate the system's robustness and resilience against the perturbance on its imagery input, we show the attack on the WAMI tracking in Figure \ref{robustness_example:a} with combined payoff $\varphi = 10.47$. The payoff is calculated as the total perturbations added to the input images for generating the current adversarial track (coloured in red) against the original track (coloured in green). If we loosen the restriction on the attacker, for example, the attack payoff is increased to $\varphi = 20.58$, with other settings remaining unchanged, we get the results in Figure \ref{robustness_example:b}. While the attacker's effort is almost doubled (from $\varphi = 10.47$ to $\varphi = 20.58$), the deviation from the original track is not increased that much. \emph{This is evidence of the system level robustness}. To have a better understanding about this, we investigate the updating process for tracking at $k = 6$ and visualise the process in Figure \ref{robustness_explain:a} and \ref{robustness_explain:b} respectively.

As shown in the Figure \ref{robustness_explain:a} and \ref{robustness_explain:b}, the blue point identifies the predicted location of the tracking vehicle, the green point is the correct observation, and other grey points are observations of other vehicles around the tracked one. Other vehicles are potential disturbances to the system. For Scene (a), the attacker makes the closest three detections invisible to the system with an attack payoff $\varphi=10.47$. For scene (b), one more observation is mis-detected with the payoff increased to $\varphi=20.58$, and thus the KF  associates the most distant observation  as the observation for updating. Nevertheless, for both scenes, the wrong observations are still within the bound, which is denoted by the dashed circle. 

We can see that the WAMI tracking system is designed in a way to be robust against the local disturbances. First, as presented in its architecture by Figure \ref{fig:detector}, the background subtraction can guarantee that only moving objects are input to the CNNs for vehicle detection; this means that error observations that can influence the tracking accuracy are discrete and finite, which are easier to control and measure than continuous errors. Second, the KF's covariance matrix leads to a search range, only within which the observations are considered. In other words, even if the attacker has infinite power -- measured as payoff -- to attack the system at some step, the possible deviations can be enumerated and constrained within a known bound.

\begin{figure}[ht] 
  \begin{subfigure}[b]{0.5\linewidth}
    \centering
    \includegraphics[width=0.9\linewidth]{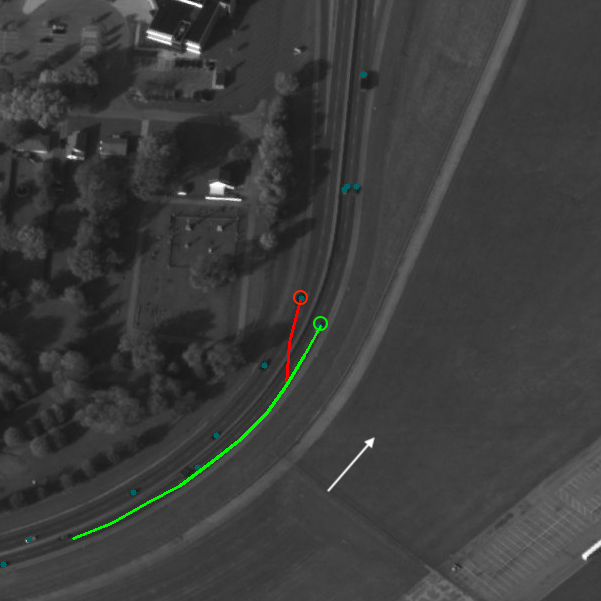} 
    \caption{vehicle tracking at $k = 9$} 
    \label{resilience_example:a} 
    \vspace{2ex}
  \end{subfigure}
  \begin{subfigure}[b]{0.5\linewidth}
    \centering
    \includegraphics[width=0.9\linewidth]{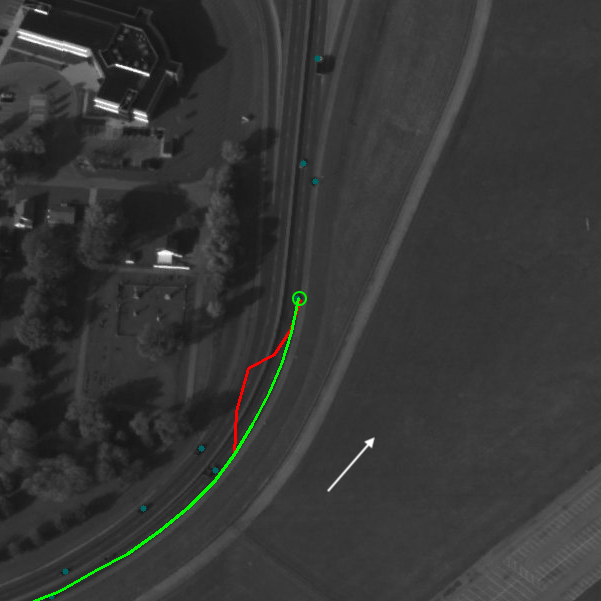} 
    \caption{vehicle tracking at $k = 12$} 
    \label{resilience_example:b} 
    \vspace{2ex}
  \end{subfigure} 
  \caption{Illustration of WAMI system's resilience to the one step attack at $k = 8$. The attack payoff $\varphi = 6.49$.}
  \label{resilience_example} 
\end{figure}

To show the system's resilience to erroneous behaviours, we consider the maximum deviation depicted in Figure \ref{resilience_example:a}, which arises from a one-step attack at $k=8$. After three steps forward, i.e., $k=12$, we have the tracking results as depicted in Figure \ref{resilience_example:b}.  Evident in this test scene, the adversarial tracking is corrected by the system itself, back to the original expected track in a short time period -- a clear evidence of resilience.

Taking a careful look into Figure \ref{resilience_example:a}, we can see that, at $k=8$, the attacked track is associated with a wrong observation in the opposite lane. Due to the consistency in the KF, this false information may disturb the original tracking, but cannot completely change some key values of the KF's state variables; for example, the direction of the velocity vector. That means that, even for the adversarial tracking, the prediction still advances in the same direction to the previous one. Hence, this wrong observation will not appear in the search range of its next step. For this reason, \emph{it is likely that the tracking can be compensated and returned to the original target vehicle}.

Another key fact is that, the KF's covariance matrix will adjust according to the detection of observations. If no observation is available within the search range, the uncertainty range -- decided by covariance matrix -- will enlarge and very likely, the error tracking can be corrected. This reflects the KF's good adaption to the errors -- another key ability in achieving resilience.

We have the following takeaway message to \textbf{RQ1}: 

\begin{framed}
The WAMI tracking system in Section \ref{sec:wami-tracker} is robust and resilient (to some extent) against the adversarial attacks on its neural network perceptional unit. 
\end{framed}

\subsection{Comparison Between Robustness and Resilience}\label{sec:expcomparison}

\begin{figure}[ht] 
  \begin{subfigure}[b]{0.5\linewidth}
    \centering
    \includegraphics[width=0.9\linewidth]{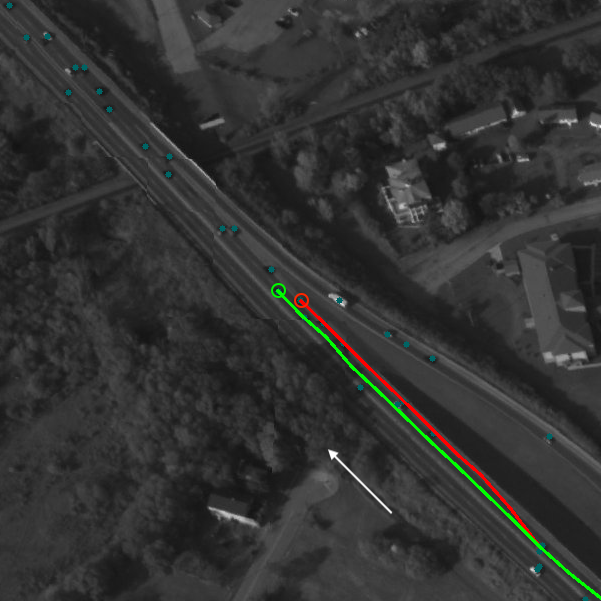} 
    \caption{robustness} 
    \label{fig:compare_robustness} 
    \vspace{2ex}
  \end{subfigure}
  \begin{subfigure}[b]{0.5\linewidth}
    \centering
    \includegraphics[width=0.9\linewidth]{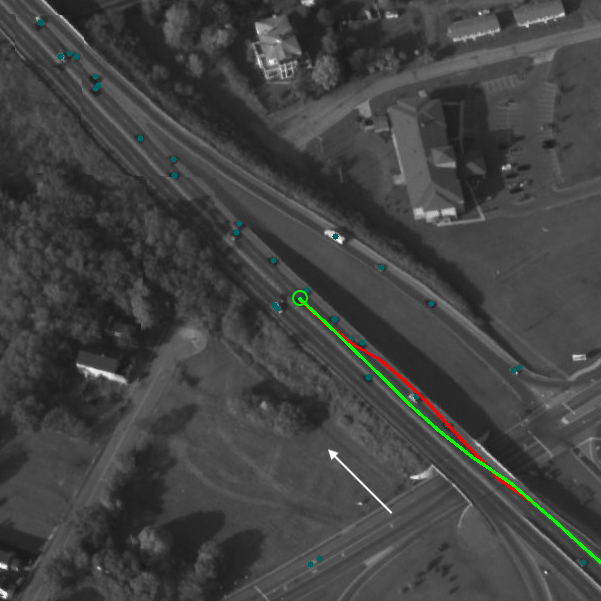} 
    \caption{resilience} 
    \label{fig:compare_resilience} 
    \vspace{2ex}
  \end{subfigure} 
  \begin{subfigure}[b]{0.5\linewidth}
    \centering
    \includegraphics[width=\linewidth]{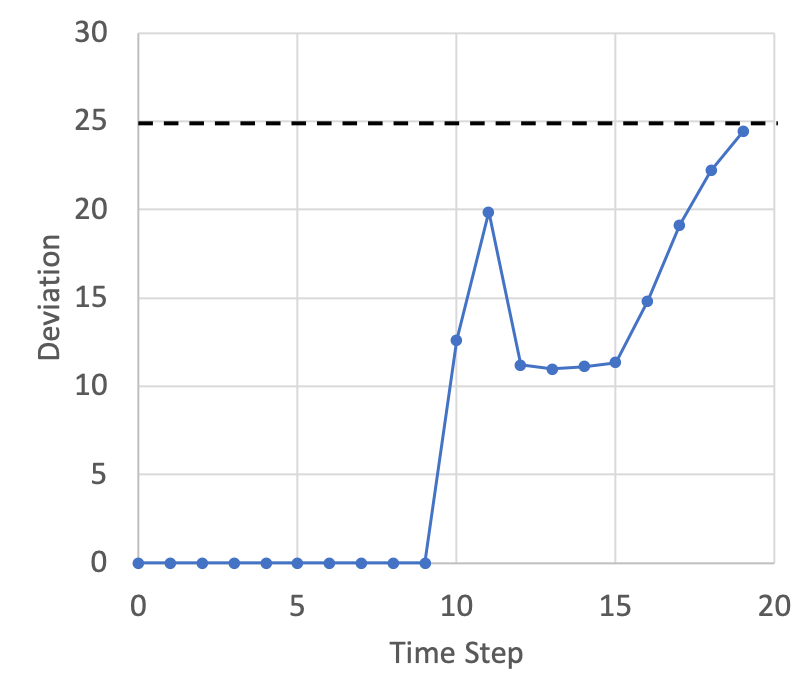} 
    \caption{deviation of robust track} 
    \label{fig:deviation_robustness} 
  \end{subfigure}
  \begin{subfigure}[b]{0.5\linewidth}
    \centering
    \includegraphics[width=\linewidth]{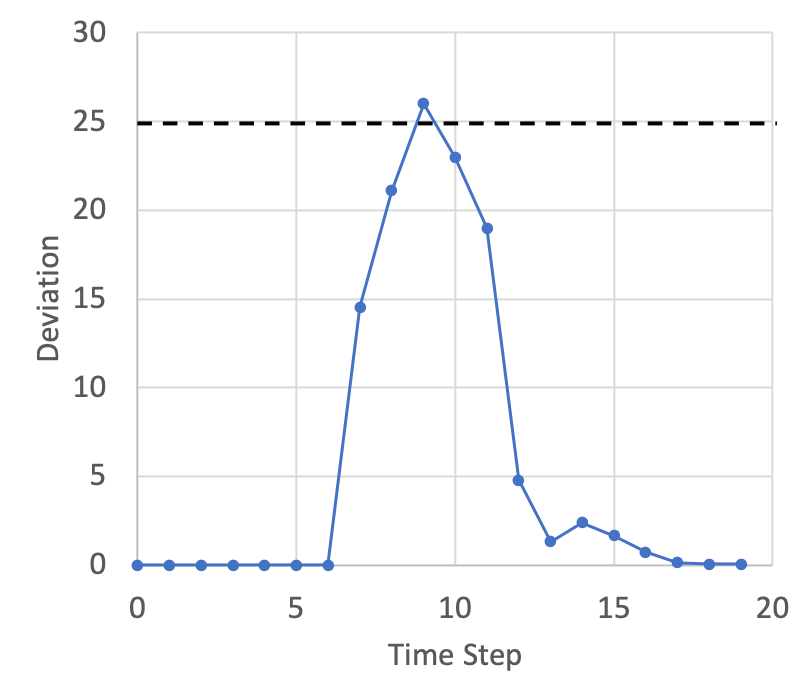} 
    \caption{deviation of resilient track} 
    \label{fig:deviation_resilience} 
  \end{subfigure} 
  \caption{Comparison between robustness and resilience in WAMI tracking}
  \label{fig:robustness_vs_resilience} 
\end{figure}

Robustness and resilience as defined in Section \ref{sec:properties} are both measures of a system's capacity to handle perturbations; however, they are not equivalent definitions. To assist in appreciating the distinction, Figure \ref{fig:robustness_vs_resilience} provides examples. 

In Figure \ref{fig:robustness_vs_resilience}, we present two typical cases to illustrate a system's robustness and resilience against the attack. Their deviations from the original tracks at each step are recorded in Figure \ref{fig:deviation_robustness} and Figure \ref{fig:deviation_resilience}, respectively. The horizontal dash line can be seen as the enhanced robustness threshold, requiring that each step's deviation is smaller than some given threshold. For the vehicle tracking in Figure \ref{fig:compare_robustness}, we can say the system satisfies the robustness property against the disturbance, since the deviation at each step is bounded. However, this tracking is not resilient to the errors due to the loss of ``recovery" property: it is apparent to see the deviation worsens over time. In contrast, for the vehicle tracking in Figure \ref{fig:compare_resilience}, the tracking is finally corrected  -- with the tracking back to the original track -- at the end even when the maximum deviation (at time step 9) exceeds the robustness threshold. Therefore, to conclude, this vehicle tracking is resilient but not robust.
 
We have the following takeaway message to \textbf{RQ2}: 

\begin{framed}
Robustness and resilience are different concepts and may complement  each other in describing the system's resistance and adaption to the malicious attack.
\end{framed}

\subsection{Quantify the Robustness and Resilience Bound}\label{sec:verificationexperiments}

In the previous subsections, we have provided several examples to show the WAMI tracking system's robustness and resilience to malicious attack. However, we still do not know to what extent the system is robust or resilient to natural environmental perturbation. In this subsection, we will show, from the verification perspective, a quantification of robustness and resilience of the WAMI tracking design in Section \ref{sec:WAMI_intro}. That is, given an original track and its associated scene, whether or not we can quantify the robustness and resilience of that track, by solving the optimisation problem defined in Equation (\ref{equ:robustnessgeneral}) and (\ref{equ:resiilencegeneral}). Moreover, as quantitative measures rather than the optimal solutions, we will further get the representative path and representative value -- as defined in Section~\ref{sec:solveoptimisation} -- of the tracking.

\begin{figure}[!ht]
        \begin{subfigure}[b]{0.5\linewidth}
                \centering
                \includegraphics[width=0.9\linewidth]{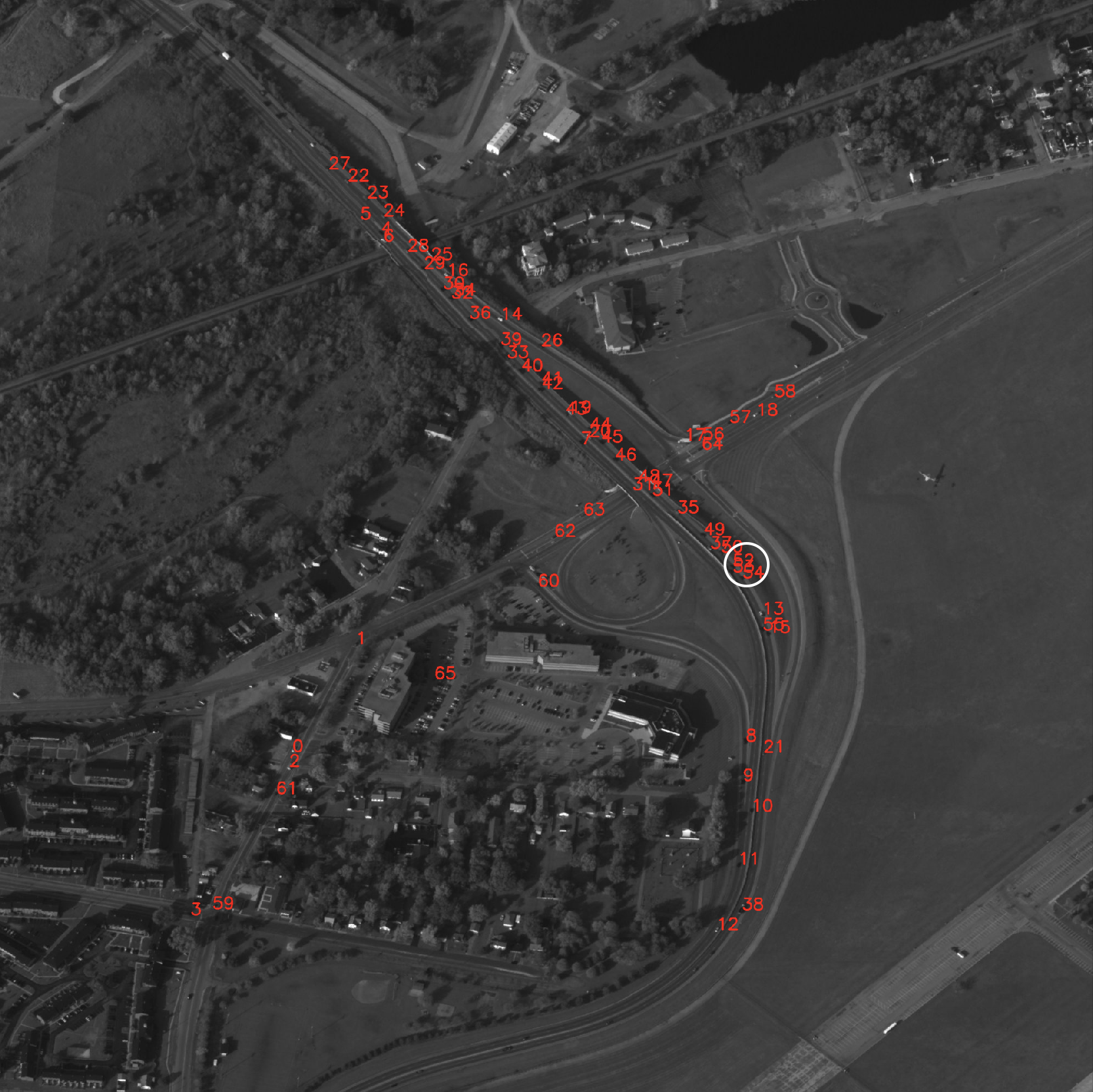}
                \caption{test scene}
                \label{fig:test_scene}
        \end{subfigure}%
        \begin{subfigure}[b]{0.5\linewidth}
                \centering
                \includegraphics[width=0.9\linewidth]{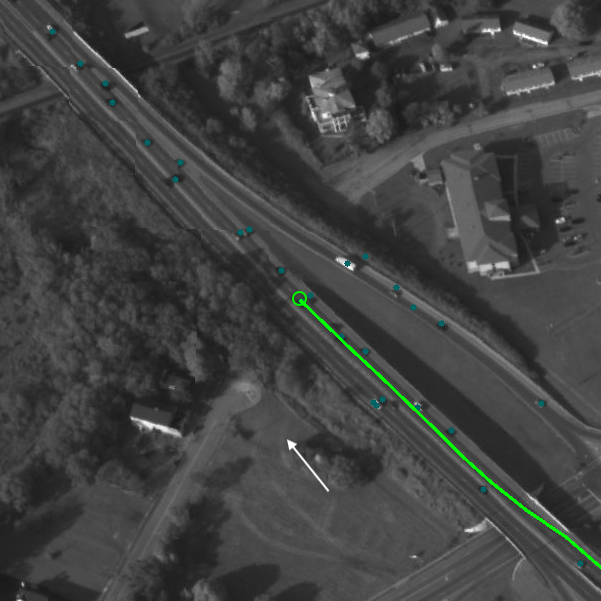}
                \caption{original track}
                \label{fig:origin_track}
        \end{subfigure}%
        \vspace{2ex}
        \begin{subfigure}{\linewidth}
                \centering
                \includegraphics[width=0.9\linewidth]{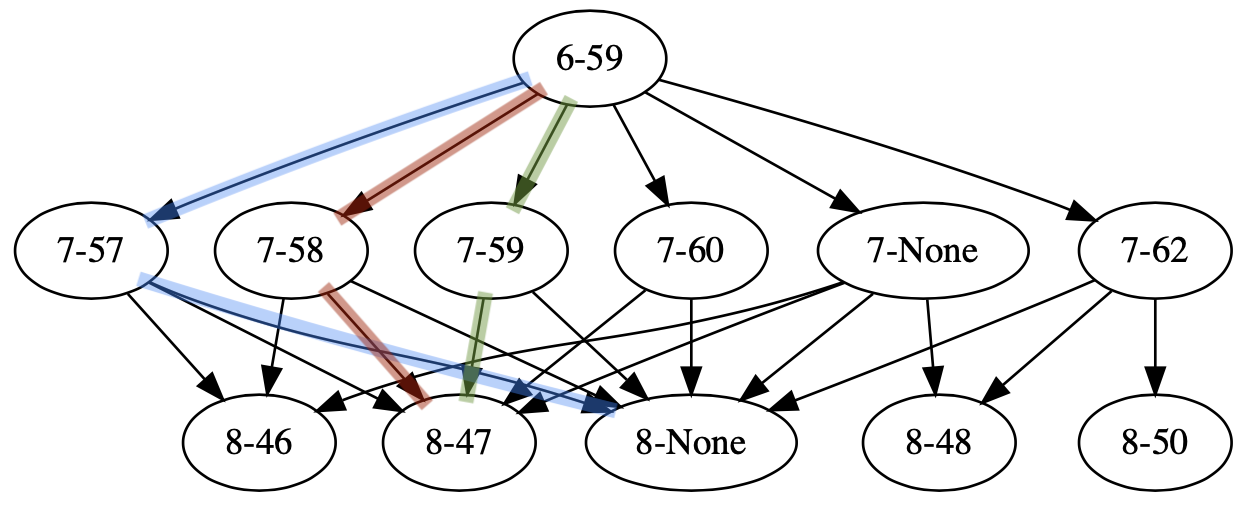} 
                \caption{Enumeration of all possible Tracks}
                \label{fig:state_space} 
        \end{subfigure}
        \caption{Attack the system at $k = 6,7,8$ on a selected scene. Tree graph exhibits all possible tracks, where green is the original track, blue is the resilience representative track, and red is the robustness representation track. The labels on the nodes represent ``(time step)-(ID of associated detection)''. }
        \label{fig:test_example}
\end{figure}

Let us consider a track and the running test scene as shown in Figure \ref{fig:test_example}. The white circle in Figure \ref{fig:test_scene} contains the target vehicle and the green line in Figure \ref{fig:origin_track} is the output of original tracking \footnote{The target vehicle in two images are in different locations due to the time difference.}. By attacking the original track from time step $k=6$ to $k=8$, we can enumerate all the possible variants (using Algorithm \ref{alg:exhausitive_search}) in Figure \ref{fig:state_space}. 

\begin{table*}[!ht]
\centering
\caption{Measures of all possible tracks for test scene in Figure \ref{fig:test_example}:
$\varphi$ is the combined payoff;
$dist^{0,e}$, $dist^{max}$ and $dist^{e,e}$ are accumulated deviation, maximum deviation and end-point deviation between the original track $\rho$ and the adversarial track $\widetilde{\rho}$ respectively.
} 
\resizebox{\textwidth}{!}
{
\begin{tabular}{cccccccccccccccccc}
\hline
Track No. & 1 & 2 & 3 & 4 & 5 & 6 & 7 & 8 & 9 & 10 & 11 & 12 & 13 & 14 & 15 & 16 & 17 \\ \hline
$\varphi$ & 20.32 & 19.44 & 12.63 & 20.15 & 20.01 & 12.56 & 13.27 & 10.62 & 9.91 & \cellcolor[HTML]{FD6864}3.10 & \cellcolor[HTML]{FFFFFF}{\color[HTML]{000000} 6.81} & \cellcolor[HTML]{67FD9A}0.00 & 18.53 & 11.72 & 12.61 & 11.65 & 11.82 \\ \hline
$dist^{0,e}$ & 529.99 & 103.17 & 7.23 & 5447.39 & 138.59 & 138.37 & 50.95 & 20.54 & 115.58 & \cellcolor[HTML]{FD6864}24.75 & \cellcolor[HTML]{FFFFFF}{\color[HTML]{000000} 530.24} & \cellcolor[HTML]{67FD9A}0.00 & 58.47 & 38.89 & 5462.11 & 5467.95 & 2430.17 \\ \hline
$dist^{max}$ & 53.65 & 26.62 & 4.42 & 936.97 & \cellcolor[HTML]{34CDF9}30.51 & \cellcolor[HTML]{FFFFFF}29.91 & 29.91 & 14.53 & 26.05 & 14.53 & 53.65 & \cellcolor[HTML]{67FD9A}0.00 & 27.44 & 22.83 & 936.97 & 936.98 & 220.95 \\ \hline
$dist^{e,e}$ & 53.65 & 0.07 & 0.01 & 936.97 & \cellcolor[HTML]{34CDF9}0.06 & \cellcolor[HTML]{FFFFFF}0.06 & 0.01 & 0.01 & 0.06 & 0.01 & 53.65 & \cellcolor[HTML]{67FD9A}0.00 & 0.01 & 0.01 & 936.97 & 936.98 & 210.13 \\ \hline
\end{tabular}
}
\label{measure_record}
\end{table*}

To find the representative value, we record all the measures for each possible track in Table \ref{measure_record}. Note that the attack payoff is calculated as the minimum perturbation for the current deviation, since we use the best attack approach, for example, Deepfool to find the shortest distance to the decision boundary in input space. Empirical parameters are set, e.g., like the robustness threshold $\epsilon_{robustness} = 120$ (i.e., the system is robust if the cumulative deviation of 20 time steps does not exceed 120), and the resilience threshold $\epsilon_{resilience} = 1$ (i.e., the system is resilient if the final deviation does not exceed 1). We remark that, these two hyper-parameters can be customised according to users' particular needs.

\begin{table}[!ht]
\centering
\caption{The outcome of robustness and resilience verification for test scene in Figure \ref{fig:test_example}: $sol_{opt}(M,\phi)$ is the optimal value; $\theta^*$ and $\rho^*$ represents the representative value and track (of the robustness and the resilience) respectively.} 
\resizebox{0.48\textwidth}{!}
{
\begin{tabular}{lll}
\hline
 & Robustness Verification & Resilience Verification \\ \hline
$sol_{opt}(M,\phi)$ & 6.81 & 53.65 \\ \hline
$\theta^*$ & 3.10 & 30.51 \\ \hline
$\rho^*$ & colored in red (Figure~\ref{fig:state_space})&  colored in blue (Figure~\ref{fig:state_space})\\ \hline
\end{tabular}
}
\label{verif_result1}
\end{table}

We then apply Algorithm \ref{alg:robustness_resilience} to search for the optimal solution to robustness and resilience verification, as defined in Equation (\ref{equ:robustnessgeneral}) and (\ref{equ:resiilencegeneral}). The verification outcome is presented in Table~\ref{verif_result1}. The results show that optimal solution $sol_{opt}(M,\phi)$ to robustness verification is $\varphi = 6.81$; the minimum attack payoff to lead to the failure of the system. The attack payoff $\varphi$ of Track No.11 and No.15 is greater or equal to $sol_{opt}(M,\phi)$, and have $dist^{0,e}$ over the robustness threshold. The result of resilience verification is $dist^{max} = 53.65$, the minimum maximum deviation from which the system cannot recover. For example, the maximum deviation $dist^{max}$ of Track No.1 and No.15 is greater or equal to $sol_{opt}(M,\phi)$, and have $dist^{e,e}$ over the resilience threshold. 

\begin{figure}[ht] 
  \begin{subfigure}[b]{0.49\linewidth}
    \centering
    \includegraphics[width=0.95\linewidth]{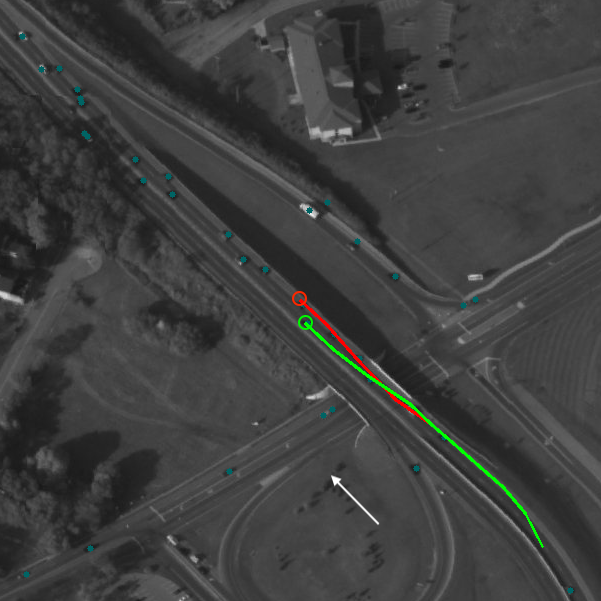} 
    \caption{robustness \\ representative track} 
    \label{fig:robust_track} 
    \vspace{2ex}
  \end{subfigure}
  \begin{subfigure}[b]{0.49\linewidth}
    \centering
    \includegraphics[width=0.95\linewidth]{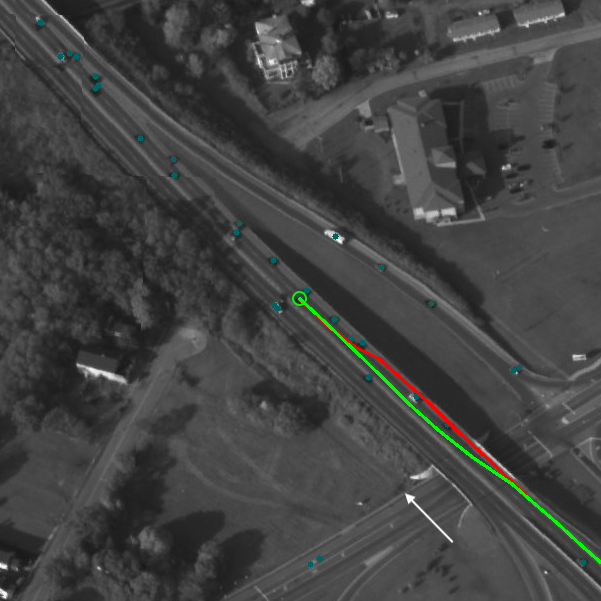} 
    \caption{resilience \\ representative track} 
    \label{fig:resilient track} 
    \vspace{2ex}
  \end{subfigure} 
  \caption{Adversarial tracks to quantify the system's robustness and resilience}
  \label{fig:output_test_scene} 
\end{figure}

Moreover, we are interested in three specific tracks: (1) the original track, denoted as Track No. 12; (2) an adversarial track to represent the system's robustness, denoted as Track No. 10; and (3) an adversarial track to represent the system's resilience, denoted as Track No. 5. We can see that, the robustness representative value of the tracking is $\varphi = 3.10$. If the attack payoff is constrained within this bound (including the bound value), the tracking of interest can have a guaranteed accumulative deviation smaller than 120. In addition, the resilience representative value is $dist^{max} = 30.51$. If the defender can monitor the tracking and control the maximum deviation to make it smaller than or equal to this bound, the system can resist the errors and recover from the misfunction. 
These two tracks, reflecting the robustness and resilience respectively, are also plotted in Figure \ref{fig:output_test_scene}.

Additionally, Table \ref{measure_record} also provides some other interesting observations that are worth discussing. For example, comparing Track No. 9, Track No.15 and Track No.16, it's not difficult to determine that the deviation is very likely to increase dramatically when disturbances go over the system's endurance. Under such circumstances, the maximum deviation normally occurs at the end of the tracking. In other words, the deviation will be increased further and further with time, and the KF is totally misled, such that it tracks other vehicles, distinctly distant to the intended one.

We have the following takeaway message to \textbf{RQ3}: 

\begin{framed}
Our verification approach can not only find the optimal values for the optimisation problems as specified in (\ref{equ:robustnessgeneral}) and Equation (\ref{equ:resiilencegeneral}), but also find representative values and paths as defined in Section~\ref{sec:solveoptimisation} to exhibit and quantify the robustness and resilience. 
\end{framed}

\subsection{Improvement to the robustness and resilience}\label{sec:expimprovement}

In this part, we apply the runtime monitor and the Joint KFs introduced in Section \ref{sec:design2}. We report whether theses two techniques can make an improvement to the WAMI tracking system through the experiments.

\begin{figure}[ht]
\begin{subfigure}{\linewidth}
\centering
\includegraphics[width=0.45\linewidth]{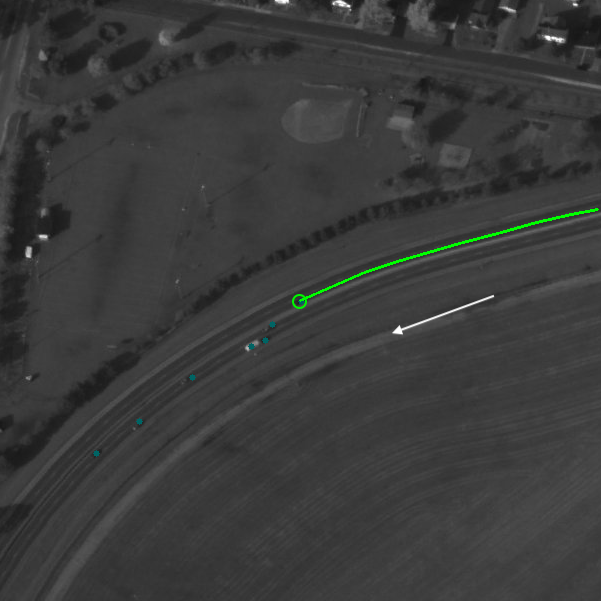} 
\includegraphics[width=0.51\linewidth]{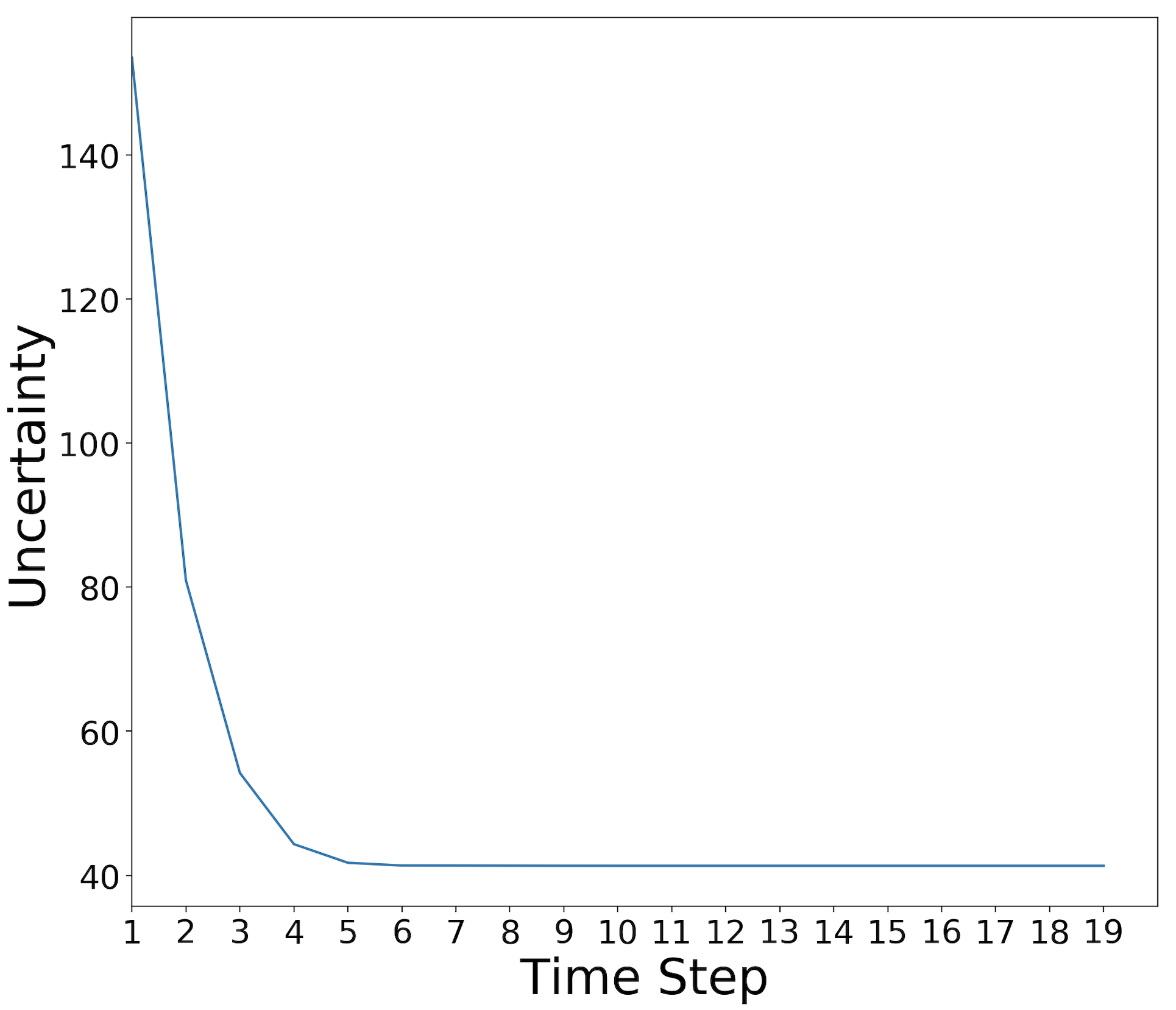} 
\caption{tracking without attack}
\label{fig:monitor_before}
\vspace{1ex}
\end{subfigure}
\begin{subfigure}{\linewidth}
\centering
\includegraphics[width=0.45\linewidth]{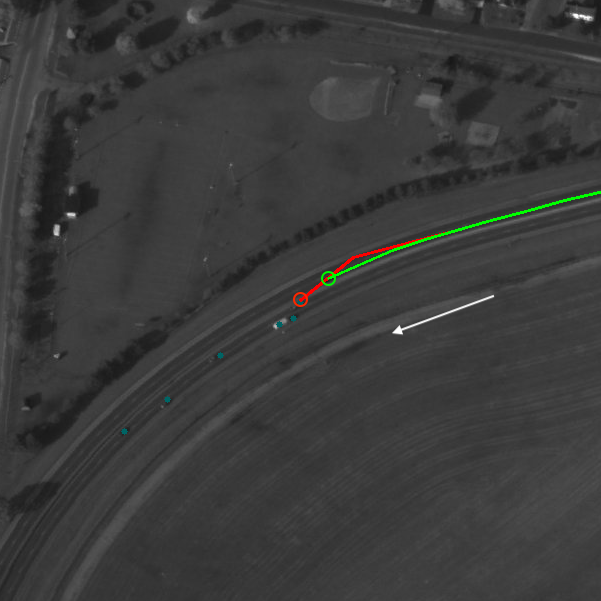} 
\includegraphics[width=0.51\linewidth]{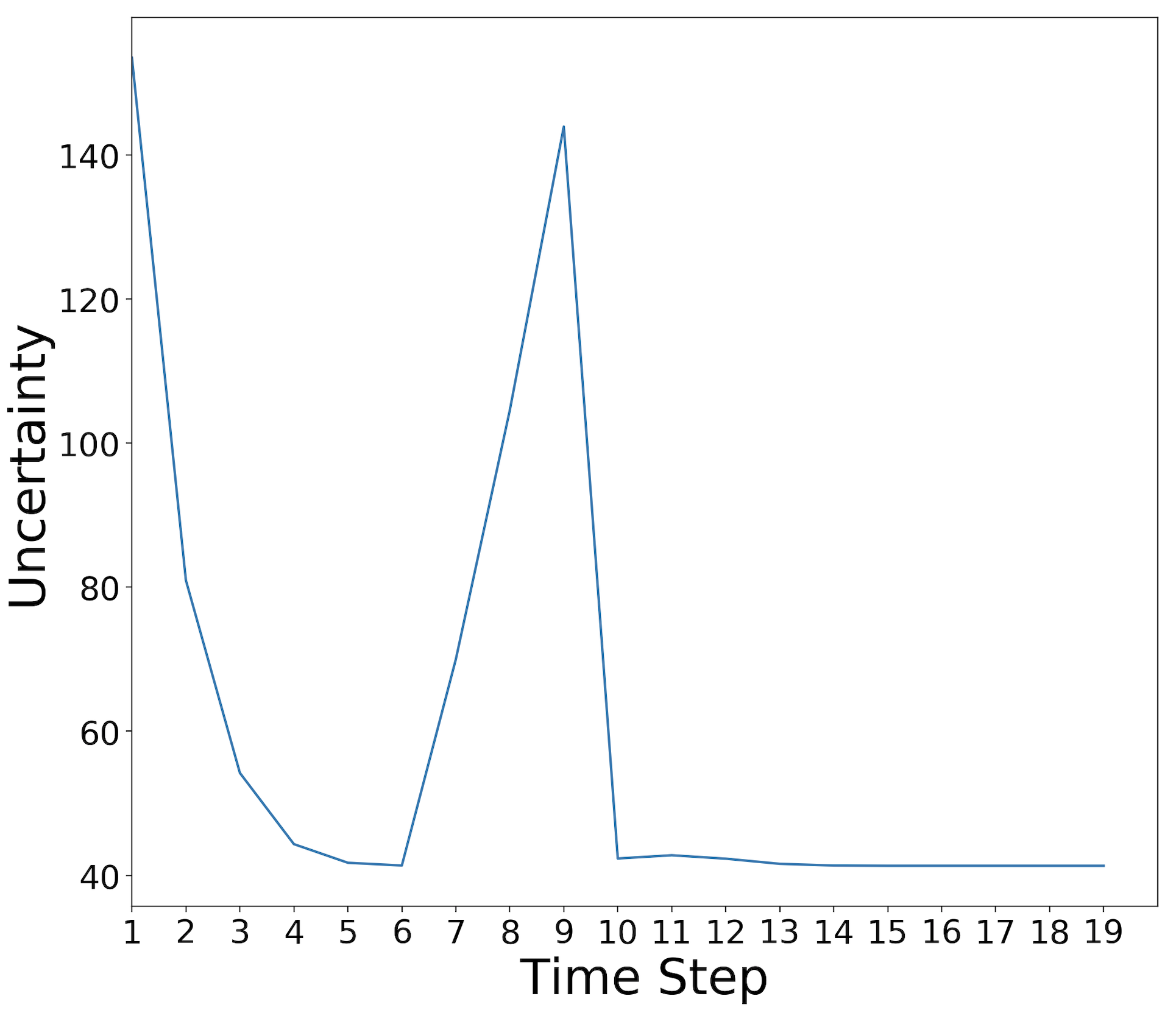} 
\caption{tracking with attack}
\label{fig:monitor_after} 
\end{subfigure}
\caption{Runtime monitoring on the WAMI system's tracking}
\label{fig:monitor}
\end{figure}

In Figure \ref{fig:monitor}, a runtime monitor runs along with 
the system in order to continuously check the Bayesian uncertainty of the KF. As discussed in Section \ref{sec:uncertaintymonitor}, a KF alternates between the prediction phase and the update phase, and the Bayesian uncertainty is gradually reduced until convergence.
This can be seen from Figure \ref{fig:monitor_before} (Right), where there is no adversarial attack on the detection component and the uncertainty curve is very smooth. However, if the system is under attack, it is likely that the expected functionality of the KF is disrupted, and the KF performs in an unstable  circumstance. Consequently, there will be time steps where no observations are available, leading to the increasing of uncertainty. When it comes to the uncertainty curve, as shown in Figure \ref{fig:monitor_after} (Right), a spike is observed.

The above discussion on uncertainty monitoring in Figure \ref{fig:monitor} is based on the condition that the environmental input is not complex and the surrounding vehicles are sparse: the mis-detection of vehicles is very likely to result in no observations seen by the system within the search range. For more complicated cases such that there are a significant number of vehicles in the input imagery, and the target tracking is more likely to be influenced by the surroundings, we need to refer to the improved design of taking joint-KFs for observation association filters in Section \ref{multi_kf}. 

\begin{figure}[ht] 
  \begin{subfigure}[b]{0.5\linewidth}
    \centering
    \includegraphics[width=0.9\linewidth]{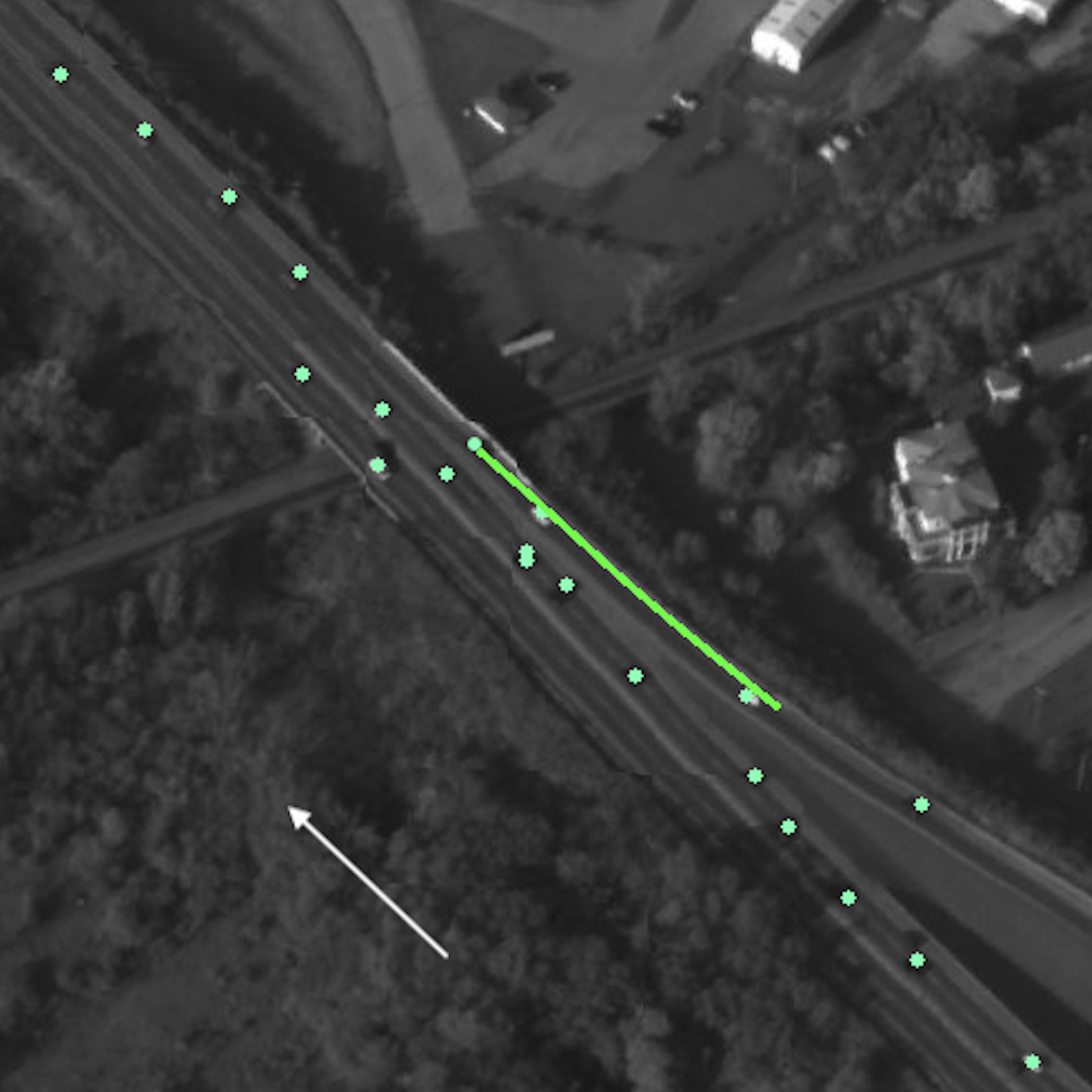} 
    \caption{tracking without attack} 
    \label{fig:multi_kf_before} 
    \vspace{2ex}
  \end{subfigure}
  \begin{subfigure}[b]{0.5\linewidth}
    \centering
    \includegraphics[width=0.9\linewidth]{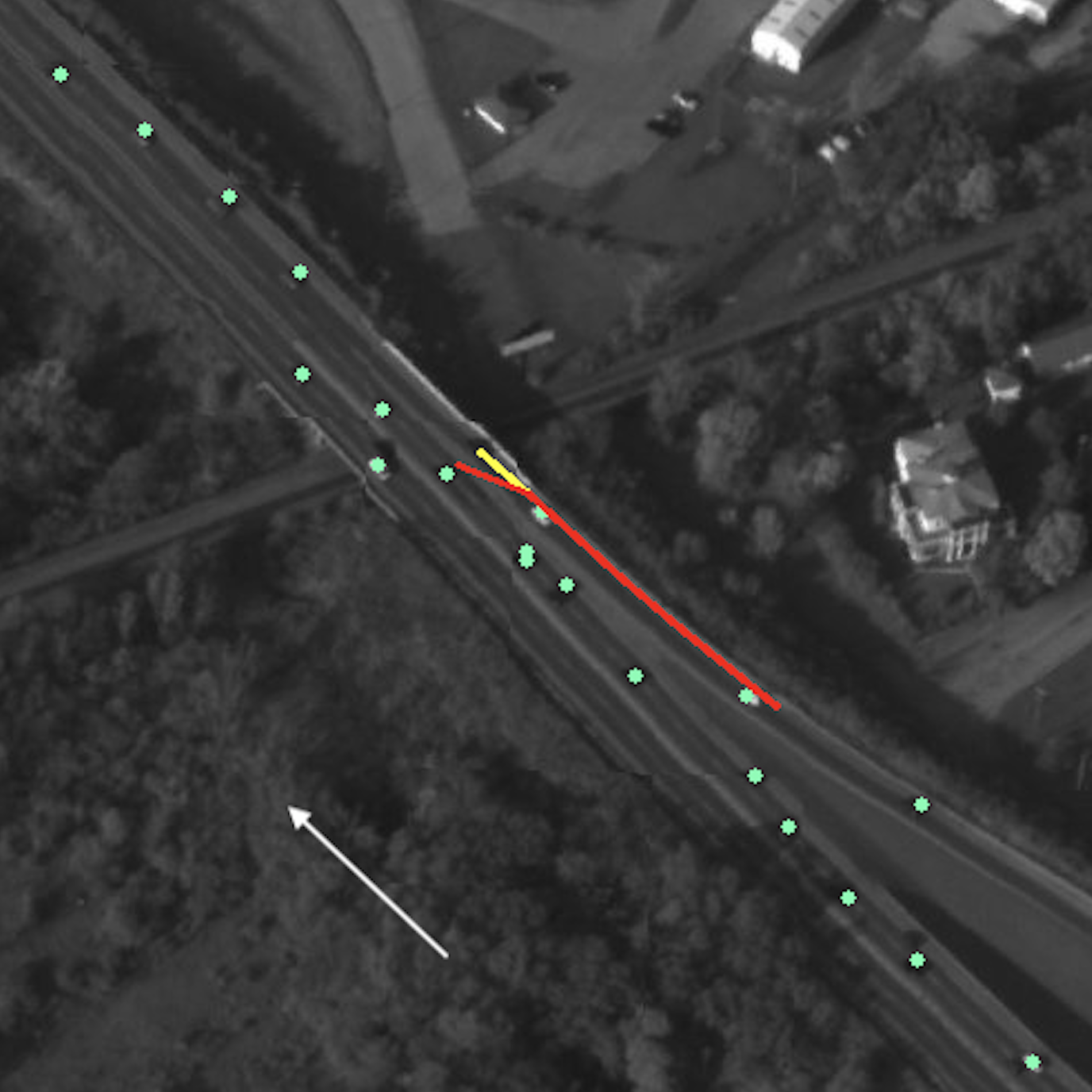} 
    \caption{tracking with attack} 
    \label{fig:multi_kf_after} 
    \vspace{2ex}
  \end{subfigure} 
  \caption{Comparison between tracking with single KF and joint KFs}
  \label{fig:multi_kf} 
\end{figure}

The main idea of implementing joint-KFs is to assign a tracking to each surrounding vehicle and the observation association is based on the maximum likelihood function as defined. In other words, if some surrounding vehicle has already been tracked by another KF, it will not be followed by the tracking of the current KF. An example is shown in Figure \ref{fig:multi_kf}.
 
In Figure \ref{fig:multi_kf_before}, the green line represents the original track and the green dots are vehicles detected by the system. Obviously, the whole system operates normally and continuously tracks the target vehicle. However, when we attack the detection component at this time step, as shown in Figure \ref{fig:multi_kf_after}, the current observation becomes invisible to the system. While using the original approach, the track is deviated to the nearest vehicle (see the red line). When the joint KFs are applied, since the surrounding vehicles are associated to other trackers, the primary tracker will not be associated with a wrong observation and will skip the update phase (and move along the yellow line). Thus, after the attack is stopped, the track is always correct and can finally be associated to the true target (as shown in Figure \ref{fig:multi_kf_after} where the yellow line overlays the green line). In the experiments, we discovered that the application of joint KFs is very effective when dealing with an attack when the vehicle traveses on a straight line, but it can be less sufficient when the attack activates while the true track is curved. This is because we adopt a constant velocity model within the dynamic model of the tracker, which is not optimal to describe the case: it makes the mean of the prediction always on a straight line and does not consider the potential direction of the movement. Therefore, when there are many detections, the data association is more likely to be wrong and lead to a larger deviation.  

\begin{table}[ht]
\centering
\caption{The outcome of robustness and resilience verification for test scene in Figure \ref{fig:test_example} with joint-KFs and Runtime Monitor (it should be read in comparison with Table \ref{verif_result1}).} 
\resizebox{0.48\textwidth}{!}
{
\begin{tabular}{lll}
\hline
 & Robustness Verification & Resilience Verification \\ \hline
$sol_{opt}(M,\phi)$ & inf. & 53.65 \\ \hline
$\theta^*$ & inf. & 30.51 \\ \hline
$\rho^*$ & none &  remain same \\ \hline
\end{tabular}
}
\label{verif_result2}
\end{table}

To understand how the above two techniques collectively improve the robustness and resilience, we consider the experiments described in Section \ref{sec:verificationexperiments}, with the improvement of using joint-KFs as data association method and the attaching a runtime monitor to the tracker that are described in Section~\ref{sec:design2}. 

\begin{figure*}[ht]
\centering
\includegraphics[width=\linewidth]{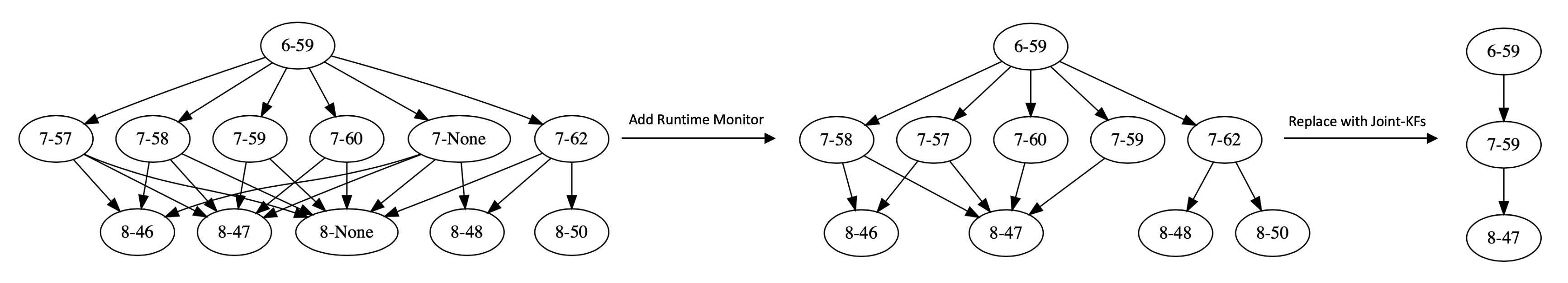} 
\caption{Improvement to WAMI tracking system by Runtime Monitor and joint-KFs (Possible adversarial tracks are reduced).}
\label{fig:robutness_improvement}
\end{figure*}

Generally speaking, after enumerating all the possible tracks, we only get the original (and true) track, with all adversarial tracks being removed. Examining Figure \ref{fig:robutness_improvement}, if we attach a runtime monitor to the system, those tracks that lack observations at some steps (detection ID is None) can be detected and eliminated. Furthermore, if we replace the system's tracking component with joint-KFs, the system can be protected from influence by incorrect observations. These two methods combined together are effective to prevent any successful attack and thereby improve the tracking robustness. As shown in Table~\ref{verif_result2}, since no adversarial track exists, the attack payoff can be theoretically infinite. In practice, the payoff is usually constrained and we can introduce an oracle to judge if the very high attack payoff is valid or not. For example, too much noise added to the image will make it unrecognized by the system. 

Regarding the resilience of the tracker, there is no evidence that the improvement methods have made a difference, even though all adversarial tracks (shown in Table \ref{measure_record}) are eliminated successfully. The reason is that when the track has switched to a wrong target (by using large $dist^{max}$, which is the case we test in determining the possibility of resilience), the history information of the true target is lost. The Kalman filter is designed to estimate the state of one object with some assumptions, and resilience to a extreme noise is beyond its capacity.

We remark that, to improve the resilience, extra component(s) that processes additional information about the target vehicle is needed, for example considering different appearances of the vehicles and the context of the road network. With this kind of knowledge, the tracking system can incorporate a way to respond to the errors, and when the track is dramatically deviated, correct itself back. This will be investigated in future work. However, what can be re-emphasised from the experiments is that the robustness and the resilience are indeed different.

We have the following takeaway message to \textbf{RQ4}: 
\begin{framed}
The runtime monitor approach can eliminate some wrong tracks, and the joint-KFs approach can reduce the risks of being influenced by wrong observations. Both of them are effective in improving the robustness, but less so for the resilience.
\end{framed}

\section{Future Work}
\label{sec:discussion}

In the experimental sections, examples have been provided to show the difference between robustness (in the context of ML) and resilience, and the preliminary strategies are investigated to quantify and improve the robustness and resilience of LE-SESs. In the following, we highlight a few aspects that are not covered in this paper but worthy for the study in the future works. All the discussions are led to the need for more comprehensive study of robustness and resilience in the context of LE-SESs, and the proposal suggested in the paper only makes a first attempt towards this.

\subsection{Monitoring Resilience}

For resilience, it is not required to always return to where the system was before the occurrence of the failure. Instead, it can resume part of its functionality. In this paper, for the LE-SESs, we define the status of ``recovered'' to be $dist^{e,e}(\rho,\widetilde{\rho})  \leq  \epsilon_{resilience}$, i.e., the distance of the final location of the attacked track is close to that of the original track, within a certain threshold.

While the status of ``recovered'' can be defined, it is harder to ``show the sign of recovering'', which is a \emph{subjective} evaluation of the system's recovering progress from an outside observer's point of view. An outside observer does not necessarily have the full details of the recovering process, or the full details of the system implementation. Instead, an observer might conduct Bayesian inference or epistemic reasoning \cite{halpenbook}  by collecting evidence of recovering. Technically, a run-time monitor can be utilised to closely monitor some measurements of the recovering process; indeed, the runtime monitor we used for Kalman filter convergence (Section~\ref{sec:uncertaintymonitor}) can be seen as a monitor of the sign of recovering. If the Bayesian uncertainty is gradually reduced and converging, 
this can be considered evidence that the system is managing the failure. On the other hand, if the Bayesian uncertainty is fluctuating, we cannot claim signs of recovery, even if it might recover in the end, i.e., satisfying the condition  $dist^{e,e}(\rho,\widetilde{\rho})  \leq  \epsilon_{resilience}$. 

\subsection{Resilience over Component Failure}

In the paper, we consider the uncertainties from the external environment of the system, or more specifically, the adversarial attacks on the inputs to the perception unit. While this might be sufficient for robustness, which quantifies the ability to deal with erroneous input, there are other uncertainties -- such as internal component failure --  that may be worthy of consideration when working with resilience, because resilience may include  not only the ability to deal with erroneous input but also the ability to cope with, and recover from, component failure, as suggested in e.g., Murray \emph{et. al} \cite{res_methods}
in software engineering. 

In LE-SESs, the component failure may include the failure of a perceptional unit or the failure of a Bayes filter. The failures of Bayes filter may include e.g., missing or perturbed values in the matrices  $\matr{F}$, $\matr{H}$, $\matr{Q}$, and $\matr{R}$. The failures of the perceptional unit may include e.g., the failure of interactions between neural networks,  the internal component failure of a neural network (e.g., some neuron does not function correctly), etc. The study of component failures, and their impact to  resilience, will be considered in our future work. 

\subsection{Robustness and Resilience on Neural Network}

Consider an end-to-end learning system where the entire system itself is a feedforward neural network -- for example a convolutional neural network as in the NVIDIA DAVE-2 self-driving car \cite{DBLP:journals/corr/BojarskiTDFFGJM16}. A feedforward network is usually regarded as an instantaneous decision making mechanism, and treated as a ``black-box''. These two assumptions mean that there is no temporal dimension to be considered. Therefore, we have that $l=0$ and $m=u=e=1$, for definitions in Equation (\ref{equ:robustnessgeneral}) and Equation (\ref{equ:resiilencegeneral}).   
Further, if the adversarial perturbation \cite{szegedy2014intriguing} is the only source of uncertainties to the system, we have 
\begin{equation}\label{equ:fnn1}
dist^{max}=dist^{1,1}(\rho,\widetilde{\rho})=\varphi^{0,1}(\widetilde{\rho})=\varphi^{l,u}(\widetilde{\rho})
\end{equation} and 
\begin{equation}\label{equ:fnn2}
\begin{aligned}
dist^{0,e}(\rho,\widetilde{\rho}) = dist^{0,1}(\rho,\widetilde{\rho}) = \\ dist^{1,1}(\rho,\widetilde{\rho}) = dist^{e,e}(\rho,\widetilde{\rho})
\end{aligned}
\end{equation} 
i.e., both the objective and the constraints of Equation (\ref{equ:robustnessgeneral}) and Equation (\ref{equ:resiilencegeneral}) are the same. 

It may be justifiable that the above-mentioned equivalence of robustness and resilience is valid because, for instantaneous decision making, both properties are focused on the resistance -- i.e., to resist the change and maintain the functionality of the system -- and less so on the adaptability -- i.e., adapt the behaviour to accommodate the change. Plainly, there is no time for recovering from the damages and showing the sign of managing the risks. Moreover, it may be that the feedforward neural network is a deterministic function, i.e., every input is assigned with a deterministic output, so there is no recovering mechanism that can be implemented. 

Nevertheless, the equivalence is somewhat counter-intuitive -- it is generally believed that robustness and resilience are related but not equivalent. We believe \emph{this contradiction may be from the assumptions of instantaneous decision making and black-box}. If we relax the assumptions, we will find that some equations -- such as $l=0$, $m=u=e=1$, and Equations (\ref{equ:fnn1}) and (\ref{equ:fnn2}) --  do not hold any more.  Actually, even for a feedforward neural network, its decision making can be seen as a sequential process, going through input layer,  hidden layers, to output layer. That is, by taking  a white-box analysis method, there is an internal temporal dimension. If so, the definitions in Equation (\ref{equ:robustnessgeneral}) and Equation (\ref{equ:resiilencegeneral}) do not equate, and capture different aspects of the feedforward network tolerating the faults, as they do for the LE-SESs.  Actually, \emph{\textbf{robustness is to ensure that the overall sequential process does not diverge, while resilience is to ensure that the hidden representation within a certain layer does not diverge}}. We remark that this robustness definition is different from that of \cite{szegedy2014intriguing}. Up to now, we are not aware of any research directly dealing with a definition of resilient neural networks.  For robustness, there is some research (such as \cite{2017arXiv170204267H,CBB2019}) suggesting that this definition  implies that of \cite{szegedy2014intriguing}, without providing a formal definition and evaluation method, as we have done.

Beyond feedforward neural networks, it will be an interesting topic to understand the similarity and difference of robustness and resilience, and how to improve them, for other categories of machine learning systems, such as deep reinforcement learning and recurrent neural networks, both of which have temporal dimension. We believe our definitions can be generalised to work with these systems.

\subsection{Aligning the Definitions}
We note here that the definition of robustness that has become generally accepted in the study of Artificial Intelligence (in particular adversarial behaviour) differs from that traditionally found in Software Engineering for critical systems where the definition relates to the systems ability to function correctly in the presence of \emph{invalid} inputs (e.g. see the IEEE standard for software vocabulary \cite{8016712}). Surprisingly, resilience is not defined therein, possibly due to the relative newness of the field. However, Murray \emph{et. al} \cite{res_methods} draw together several sources to suggest that resilient software should have the capacity to withstand and recover from the failure of a critical component in a timely manner. This ties in with our definition in Subsection~\ref{def_res}, but can be considered as narrower since it does not extend to attributes that may prevent component failure (as in dealing with external perturbations). 
To facilitate the move toward integrating ML technologies in high-integrity software engineer practices, the definitions currently being adopted in state-of-the-art ML research, such as robustness and resilience, should be aligned to existing and accepted software engineering definitions.

\section{Related Work}\label{sec:related} 

Below, we review research relevant to the verification of robustness and resilience for learning-enabled systems. 

\subsection{Safety Analysis of Learning Enabled Systems}\label{sec:relatedsafety}

ML techniques have been confirmed to have potential safety risks \cite{GSS2014}. Currently, most safety verification and validation work focuses on the ML components, including formal verification  \cite{HKWW2017,katz2017reluplex,xiang2017output,GMDTCV2018,LM2017,wicker2018feature,RHK2018,wu2018game,ruan2018global,LLYCH2018} and 
coverage-guided testing \cite{sun2018concolic,PCYJ2017,sun2018testing,ma2018deepgauge,sun2018concolicb,huang2021coverage,nollerhydiff}. Please refer to  \cite{huang2020survey} for a recent survey on the progress of this area.

Research is sparse at the system level, and there is none (apparent) on LE-SESs. In \cite{dreossi2019compositional}, a compositional framework is developed for the falsification of temporal logic properties of cyber-physical systems with ML components. Their approaches are applied to an Automatic Emergency Braking System. A simulation based approach \cite{tuncali2018reasoning} is suggested to verify the barrier certificates -- representing safety invariants -- of autonomous driving systems with an SMT solver. In both papers, the interaction -- or synchronisation -- between ML and other components is through a shared value, which is drastically different from the neural network enabled state estimation, where the synchronisation is closer to the message-passing regime.  
Moreover, the erroneous behaviours and the specifications of this paper are different from those of \cite{dreossi2019compositional,tuncali2018reasoning}. These differences suggest that the existing approaches cannot be extended to work with our problem.

In addition, in \cite{10.1145/3302504.3311806}, a system with a sigmoid-based neural network as the controller is transformed into a hybrid system, on which the verification can be solved with existing tools. This approach may not generalise to general neural networks since it heavily relies on the fact that the sigmoid is the solution to a quadratic differential equation. In \cite{10.1145/3302504.3311814}, a gray-box testing approach is proposed for systems with learning-based controllers, where a gradient based method is taken to search the input space. This approach is heuristic, and based on the assumption that the system is differentiable. The LE-SESs cannot be verified with these approaches. 

Moreover, Several research studies have explored the robustness of SES in the face of false information injection into sensors. In these studies, false information is typically modeled as Gaussian noise~\cite{niu2012system,yang2016false,song2023finite}. Additionally, there are the Denial-of-Service (DoS) attacks, which are particularly damaging cyber threats that can block transmission channels and hinder the network from functioning normally~\cite{song2023switching}. Another area of concern is model uncertainty, which often arises due to equipment aging~\cite{zhou2022robust}.

To address these challenges, advanced learning control algorithms have been developed. For instance, an adaptive neural finite-time resilient dynamic surface control (DSC) strategy ensures that all states of the closed-loop system remain bounded. This strategy aims for the stabilization errors of each subsystem to converge to a minimal region in a finite timeframe. As detailed in \cite{song2023switching}, a switching-like event-triggered state estimation strategy has been developed specifically for Reaction-Diffusion Neural Networks, especially when subjected to DoS attacks. Furthermore, \cite{zhou2022robust} introduces a robust point-to-point iterative learning control framework, designed to tackle energy problems that are influenced by both model uncertainty and system disturbances.

This paper distinguishes itself from existing research in that we consider adversarial attacks on the learning component, which furnishes observations for the state estimation. To enhance robustness, we further employ a runtime monitor and a joint-KF.

\subsection{Robustness and Resilience}

The endeavor to define the two perspectives -- robustness and resilience -- regarding a system's predictable operational outcome is explored from various angles. For instance, studies in automation \cite{rieger2014resilient, francis2014metric} closely associate robustness with a predefined purpose, characterized by certain values and judgments. Criteria like "correctness" and "validity" are then imposed to evaluate a system or component. This perspective, complemented by others that focus on safety and risk aversion as outlined in Section~\ref{sec:relatedsafety}, highlights the extent to which systems adhere to predefined criteria, even in the face of deviations. However, discussions seldom touch upon the counterpart of this concept: resilience.

Resilience encapsulates a distinct set of characteristics, particularly seen in social science systems. These include inherent openness, multi-dimensionality, adaptive accommodation, and diversity in both value and evolution. These under-explored traits warrant further attention to facilitate a comprehensive understanding of resilience in AI. Notably, before these terms became prevalent in representative multidisciplinary studies, such as system and environmental research \cite{dale2010community}, there were discussions on the methodologies concerning robustness and resilience. In \cite{read2005some}, the exploration revolves around robustness and resilience in social systems by examining human societies to elucidate human behavior. It was further discerned that the self-monitoring of a system's state, which aligns with our enhancement techniques through runtime monitoring, plays a pivotal role in both robustness and resilience. \cite{scholz2012risk} compares four system properties -- risk, vulnerability, robustness, and resilience -- from a decision-theoretical standpoint, proposing that robustness opposes (static) vulnerability, while resilience parallels (dynamic) vulnerability when considering the known threats or hazards a system faces. This perspective aligns closely with our definitions. Beyond these methodological deliberations, this paper showcases the difference between robustness and resilience in a learning-enabled autonomous system and formally verifies these properties.

While these broad considerations of robustness and resilience are invaluable, we must reiterate an earlier observation: the definitions used in ML applications differ from those embraced by the broader software engineering community, including the IEEE standards. This discrepancy warrants resolution.

\section{Conclusion}\label{sec:concl}

This paper introduces a formal verification-guided approach for the design of learning-enabled state estimation systems. While our initial design of the state estimation system performs commendably in its tracking task, its robustness and resilience show vulnerabilities. Our formal verification approach identifies and addresses these weaknesses, leading to an enhanced system design with improved robustness.

The research presented in this paper lays the groundwork for several subsequent research activities. Firstly, although our definitions of robustness and resilience are tailored for LE-SESs, a detailed examination is necessary to determine if, and how, these definitions can be generalized to a wider range of learning-enabled autonomous systems. Another  attempt \cite{BENSALEM2024100941} towards this has been made, where a special type of probabilistic laballed transition systems are used for the modelling of LE-SESs, although no  empirical studies have been conducted. Secondly, runtime verification techniques warrant additional exploration. The formal verification technique is determined to be NP-complete for robustness. Our verification algorithm is effective for the WAMI tracking system because we conduct offline analysis. A more streamlined runtime verification technique will be invaluable when dealing with large-scale, networked systems comprising hundreds or thousands of components, as also suggested in \cite{10.1007/978-3-031-46002-9_4}. Lastly, there's a need for diverse strategies to enhance robustness and resilience. We have delved into two strategies: (1) the benefit of integrating collaborative components and (2) the value of a runtime monitor. Investigating other strategies and comparing their effectiveness will be a captivating avenue for future research. 

\section*{Acknowledgment}
This work is supported by the UK EPSRC projects on Offshore Robotics for Certification of Assets (ORCA) [EP/R026173/1], End-to-End Conceptual Guarding of Neural Architectures [EP/T026995/1], the UK Dstl projects on Test Coverage Metrics for Artificial Intelligence, and the National Key Research and Development Program of China, "Dynamically Scalable Mimic Computing Systems and Construction Methods" (Project No. 2022YFB4500900).

\bibliographystyle{plain}
\bibliography{ref}
\end{document}